%% file: Heyne_2026_Multimodal_Approaches_for_Visually-Rich_Document_Type_Classification.tex
\documentclass[11pt]{scrartcl}

\usepackage[T1]{fontenc}
\usepackage{lmodern}
\usepackage[utf8]{inputenc}
\usepackage[english]{babel}

\usepackage[a4paper,margin=2.5cm]{geometry}
\usepackage{microtype}

\usepackage{graphicx}
\usepackage{tikz}
\usetikzlibrary{positioning}

\usepackage{booktabs}
\usepackage{arydshln}
\usepackage{listings}
\usepackage{fvextra}
\usepackage{rotating}
\usepackage{caption}

\usepackage{hyperref}
\usepackage{comment}
\usepackage{enumerate}

\usepackage[draft]{todonotes}

\usepackage{glossaries}
\glsdisablehyper

\newacronym{VRD}{VRD}{Visually Rich Document}
\newacronym{VRDU}{VRDU}{Visually Rich Document Understanding}
\newacronym{SOTA}{SOTA}{state-of-the-art}
\newacronym{LLM}{LLM}{Large Language Model}
\newacronym{CNN}{CNN}{Convolutional Neural Network}
\newacronym{RNN}{RNN}{Recurrent Neural Network}
\newacronym{MLM}{MLM}{Masked Language Modeling}
\newacronym{OCR}{OCR}{Optical Character Recognition}
\newacronym{MLLM}{MLLM}{Multimodal Large Language Model}
\newacronym{FT}{FT}{fine-tuning}
\newacronym{MoE}{MoE}{Mixture-of-Experts}
\newacronym{MLP}{MLP}{Multi-Layer Perceptron}
\newacronym{NLP}{NLP}{Natural Language Processing}

\title{Multimodal Approaches for Visually-Rich Document Type Classification: A Comparative Analysis}
\usepackage{orcidlink}
\author{
	\begin{tabular}{r l}
		Catyana Heyne\,\orcidlink{0009-0002-3807-012X} &
		{\small\textit{(catyana.heyne@oth-regensburg.de)}}\\
		Jürgen Frikel\,\orcidlink{0000-0002-2391-9859} &
		{\small\textit{(juergen.frikel@oth-regensburg.de)}} \\
		Filippo Riccio\,\orcidlink{0009-0001-3005-2704}&
		{\small\textit{(filippo.riccio@oth-regensburg.de)}}
	\end{tabular}\\[2.5em]
	\textit{OTH Regensburg, Germany}
}

\date{}
\usepackage{scrlayer-scrpage}
\ihead{Multimodal Approaches for Visually-Rich Document Type Classification}
\ohead{Heyne et al.}

\begin{document}
	
	\maketitle
	\thispagestyle{empty}
	
	\begin{abstract}
		Document type classification in visually rich documents remains challenging, as relevant information is distributed across textual, visual, and layout modalities. To capture this complexity, current approaches rely on diverse multimodal modeling strategies, resulting in heterogeneous architectures that complicate systematic comparison. 
		This variability is also reflected in existing comparative studies, which often rely on heterogeneous evaluation setups, further complicating systematic comparison and making it difficult to assess progress.
		
		To address these limitations, this work provides a structured analysis of multimodal design strategies across transformer- and LLM-based architectures, combined with a controlled empirical comparison within a unified experimental framework. Specifically, four representative models (LayoutLMv3, Donut, Qwen3-VL-32B-Instruct, and Qwen3-32B) are evaluated on the RVL-CDIP benchmark to systematically analyze the contributions of text, image, and layout information for document type classification, with a particular focus on contrasting OCR-dependent and OCR-free approaches.
		
		The results show that specialized multimodal Transformers outperform LLM‑based approaches on visually rich and layout‑intensive documents. Image information contributes most strongly to reliable classification, while OCR‑derived text provides useful but secondary support. These findings highlight that multimodal processing remains essential for documents with pronounced layout structure.
		Overall, the study provides a systematic basis for comparing multimodal architectures and offers practical guidance for selecting effective feature combinations and model designs for document type classification.
		\medskip\medskip
		
	\end{abstract}
	
	\include{files/introduction}
	\include{files/literature}
	\include{files/methodology}
	\include{files/experimental_setup}
	\include{files/results}
	\include{files/discussion}
	\include{files/conclusion}
	
	\bibliographystyle{IEEEtran}
	\bibliography{literature.bib}

\end{document}

%% file: files/introduction.tex
\section{Introduction}

\paragraph{Motivation and Problem Statement} 
Across enterprise and public-sector workflows, large volumes of business documents circulate daily, including forms, invoices, emails, reports, and regulatory notices. 
These documents play a central role in business and administrative processes, making accurate classification at the point of entry essential for ensuring consistent and reliable processing.
Automated type assessment can replace manual intake inspection, support structured prioritization, and enable compliance‑oriented archiving. 
In addition, it reduces repetitive, low-value manual tasks and enables downstream processing steps, such as automated information extraction based on the identified document type.

The difficulty of document type classification is closely tied to the characteristics of the documents themselves. 
In practice, most operational business documents belong to the class of \glspl{VRD}, which encode information jointly through textual content, layout organization, typographic cues, graphical elements (e.g., images, tables, icons, stamps), form structures, and semantically coherent regions.
Consequently, \gls{VRDU} becomes a prerequisite for reliable document processing: downstream tasks such as document type classification, information extraction or document-centric question answering can only be performed effectively when the system first understands the document’s structural and visual context.

\paragraph{Existing Approaches}
The effectiveness of models for document type classification in \gls{VRDU} is fundamentally determined by how they encode and process textual, visual, and spatial information. 
In line with observations from recent \gls{VRDU} surveys~\cite{Ding.2025,Ding.2026}, many early deep learning-based \gls{VRDU} systems handled textual and visual information in separate model components:
\gls{OCR}-derived text was modeled as linear token sequences using text-centric sequence encoders (e.g., \glspl{RNN}), while visual and layout information was handled independently using primarily \gls{CNN} feature extractors~\cite{LeipengHao.2016}, including region-based variants such as R-CNN-style detectors~\cite{He.2017,Ren.2015}. 

Later work employed unimodal, text-only Transformer encoders (e.g., BERT-based models~\cite{Devlin.2019,Liu.2019}) as baselines by applying self-attention to linearized \gls{OCR}-extracted text~\cite{Rombach.2025}. 
Although textual and visual features were sometimes combined through late fusion or heuristic rules, their representations were largely learned independently, resulting in limited cross-modal interaction and restricting the modeling of spatially grounded semantics~\cite{Ding.2025,Ding.2026}.

To address these limitations, multimodal Transformer architectures were introduced that use attention to fuse textual tokens, visual and spatial/layout features in a unified representation, enabling coherent cross-modal reasoning on visually rich documents (e.g., LayoutLMv3~\cite{Huang.2022}, which integrates OCR-derived text, 2D layout coordinates, and image patches within a multimodal encoder; and Donut~\cite{Kim.2021}, an OCR-free image-to-text framework based on a vision encoder and an autoregressive decoder).

Building on these developments, recent general-purpose models based on \glspl{LLM}~\cite{Bai.2025,Hong.2025} combine vision encoders with large language models, leveraging large‑scale pretraining and instruction tuning to achieve strong zero‑shot and few‑shot performance. Despite this progress, multimodal \glspl{LLM} still tend to lag behind specialized, fine-tuned Transformer models on domain-specific \gls{VRDU} tasks, and often struggle to generalize across heterogeneous real-world document formats~\cite{Ding.2025}.

\paragraph{Research Gap}
Administrative and business documents exhibit substantial structural variability, ranging from highly diverse layouts and formatting across sources to more standardized templates~\cite{Gbada.2025}. 
This heterogeneity is further reflected in the diversity of model architectures applied to document understanding tasks, ranging from OCR-based transformers to OCR-free, general-purpose multimodal \glspl{LLM}, complicating direct performance comparisons.
Several papers illustrate these challenges:
While existing literature reviews~\cite{Ding.2025,Ding.2026,Rombach.2025,Sassioui.2023} provide comprehensive architectural overviews and report aggregate measures such as overall accuracy to support model comparisons, these evaluations remain limited, as they lack the fine-grained empirical analyses required for rigorous and truly comparable evaluation.
Rombach and Fettke~\cite{Rombach.2025} further highlight that variations in evaluation design and insufficient methodological reporting severely limit cross-study comparability. As a result, it becomes difficult to assess whether reported performance gains truly reflect methodological advances. Similarly, Scius-Bertrand et al.~\cite{SciusBertrand.2024} present a small-scale comparison on food label images in which selected OCR-based Transformer models outperform specific OCR-free architectures.
However, the limited scope of this study restricts the generalizability of its findings.
Compounding these issues, the field continues to lack well-established and broadly applicable benchmarking frameworks for Document Understanding and especially for Type Classification~\cite{Borchmann.2021,Larson.2023}. Consequently, there remains a clear need for fine-granular, standardized, and methodologically rigorous evaluation protocols to reliably assess modern multimodal Transformer- and LLM-based models.

\paragraph{Objectives}
To address this research gap, this work examines multimodal approaches for document type classification with a focus on improving the comparability of these heterogeneous model architectures. It combines a structured review of existing methods with a controlled empirical analysis of representative models, enabling a consistent comparison between specialized Transformer approaches and general-purpose, LLM-based multimodal systems. Particular attention is given to multimodal design choices, especially the contrast between OCR-dependent and OCR-free architectures, and to how these choices influence performance and model limitations.

The empirical study presented in this work evaluates four representative architectures: LayoutLMv3~\cite{Huang.2022} (OCR‑dependent Transformer), Donut~\cite{Kim.2021} (OCR‑free Transformer), Qwen3‑VL-32B-Instruct~\cite{Bai.2025} (OCR-free LLM) and Qwen3-32B~\cite{Yang.2025} (OCR-dependent LLM) within a harmonized pipeline that standardizes inference configurations. This unified setup enables a focused multimodal assessment on the widely recognized Document Type Classification benchmark RVL‑CDIP~\cite{Harley.2015}.

\paragraph{Contributions}
The contributions of this work are summarized as follows: 
\begin{enumerate}
	\item \textbf{Structured review of multimodal \gls{VRDU} architectures.}
	We summarize how existing Transformer- and LLM-based approaches combine visual, textual, and layout information for Document Type Classification and highlight the key design differences between OCR-dependent and OCR-free multimodal architectures.
	\item \textbf{Controlled comparison of representative multimodal models.}
	We address the lack of consistent evaluation setups in multimodal VRDU by comparing four representative architectures (LayoutLMv3, Donut, Qwen3‑VL-32B-Instruct and Qwen3-32B) within a unified pipeline using the RVL‑CDIP document type classification benchmark. This setup enables a fair assessment of their respective strengths, weaknesses, and performance characteristics.
	\item \textbf{Empirical assessment of current multimodal capabilities.}
	We provide a clear, empirically grounded characterization of how current multimodal Transformer- and LLM-based models perform in Document Type Classification and derive the key trade-offs between OCR-dependent and OCR-free architectures, highlighting their practical strengths and limitations.
\end{enumerate}

\paragraph{Results}
The findings are consistent with the common assumption~\cite{Ding.2026} that multimodal approaches are essential for reliable document type classification on visually rich and layout-intensive documents. Across all experiments, specialized multimodal Transformer models consistently outperform general-purpose \glspl{LLM}, underscoring the benefits of architectures tailored to document understanding tasks. The modality-focused analysis shows that visual features are the primary driver of robust classification performance, while OCR-derived text mainly provides complementary support. OCR-free models preserve visual and layout information by operating directly on document images and avoid the additional overhead associated with external OCR frameworks. In contrast, approaches that rely exclusively on OCR-derived text exhibit clear limitations when applied to documents with complex layout structures. Overall, the findings confirm that multimodal, layout-aware architectures are essential for document type classification on layout-intensive documents, whereas documents with predominantly linear and visually simple text structures benefit to a lesser extent from multimodal modeling. The results show that improving classification performance often comes at the cost of reduced deployment flexibility or increased model specialization.

\paragraph{Paper Structure}
Section~\ref{sec:overview} introduces the scientific background, focusing on architectural approaches to document type classification, alongside an analysis of recently prevalent model architecture paradigms and commonly used type classification datasets.
Section~\ref{subsec:methodology} outlines the study design and motivates the selection of representative model classes. Section~\ref{sec:experimental_setup} details the experimental setup, including dataset, evaluation metrics, hardware, software and model‑specific configurations. Section~\ref{sec:results} presents the evaluation results, with emphasis on class‑wise model performance and practical usability. Section~\ref{sec:discussion} discusses the findings in a broader context, including the interpretation of multimodal effects, model adaptation strategies, inference efficiency and dataset considerations and section~\ref{sec:conclusion} concludes the report.

%% file: files/literature.tex
\section{Overview of Existing  Approaches}\label{sec:overview}
Visual‑rich documents (\Glspl{VRD}) refer to document types in which the overall information content arises not solely from textual components, but from the interplay of visual, spatial and semantic relationships among document elements~\cite{Gbada.2025}. 
Typical examples include business documents such as forms, invoices, reports and tables, which rely heavily on layout structure and the arrangement of textual and visual cues.  
To address this challenge in automated document type classification, a variety of computational approaches have been proposed, differing substantially in architectural design and modality integration.
This section provides an overview of these approaches and discusses their respective strengths and limitations.

\subsection{Evolution of Architectural Approaches}

\subsubsection{Conventional Deep Learning Methods} 
Due to their semi-structured or unstructured nature, \glspl{VRD} pose substantial challenges for automated document type classification. Their heterogeneous and highly variable layouts, together with the complex interplay of textual, visual and spatial information, limit the applicability of conventional document analysis methods.

Many early deep learning approaches adopted modular architectures in which the different input modalities were processed separately~\cite{Ding.2025,Ding.2026}. \gls{OCR}-extracted text was typically treated as a one-dimensional sequence and encoded with text-focused sequence models (e.g., RNNs), whereas visual appearance and layout cues were captured independently using CNN-based spatial feature extractors~\cite{LeipengHao.2016}, including region-oriented detectors such as R-CNN variants~\cite{He.2017,Ren.2015}. Even when these modality-specific features were aggregated at later stages, the lack of jointly learned representations limited cross-modal interaction and hindered the modeling of spatially grounded semantics essential for understanding visually rich documents \cite{Ding.2025,Ding.2026}.

\subsubsection{Multimodal Transformer Models} 
Recent multimodal Transformer architectures directly address the limitations of conventional methods by enabling the joint processing of heterogeneous data streams. State-of-the-art models such as DocFormerv2~\cite{Appalaraju.2024} and LayoutLMv3~\cite{Huang.2022} incorporate not only textual input but also visual and spatial features, which is reported to support more coherent and robust cross‑modal reasoning~\cite{Ding.2026}. Through their self-attention mechanism, these models learn associations between textual tokens, visual regions, and spatial layout structure, allowing them to capture cross‑modal relationships that conventional architectures typically fail to model.

Pretraining on large-scale \gls{VRD} corpora, such as the IDL collection~\cite{IDL.2026} (13M documents~\cite{Appalaraju.2024}) and IIT-CDIP~\cite{Lewis.2006}, combined with multimodal pretraining objectives such as masked language modeling~\cite{Huang.2022}, has been shown to improve the models' ability to capture cross-modal dependencies~\cite{Ding.2025, Rombach.2025}. The resulting multimodal base models can subsequently be fine-tuned for a wide range of downstream tasks, including document type classification, information extraction, and layout analysis.
While such specialized models are reported to achieve strong performance on widely used benchmarking datasets such as CORD~\cite{SeunghyunPark.2019}, FUNSD~\cite{Jaume.2019} or RVL-CDIP~\cite{Harley.2015}, they remain constrained by their design. They lack genuine logical reasoning, cannot infer information that is not explicitly present in the document, and generalize poorly to previously unseen document formats or templates~\cite{Bhattacharyya.2025}. 
Moreover, these gains tend to diminish under domain shift. Fine-tuning typically relies on large, carefully curated and annotated corpora, which often fail to fully capture real-world layout variability. In addition, the high computational demands of fine-tuning further hinder adaptation in low-resource settings~\cite{Ding.2025,Rombach.2025}.

\subsubsection{Multimodal LLM Models}
Driven by the rapid progress of generative LLMs, the field of \gls{VRDU} has expanded considerably since the emergence of LLM-based \gls{VRD} approaches in 2023, as evidenced by the increasing number of publications (from 2 in 2023 to 9 in 2024)~\cite{Rombach.2025}.

Recent LLM‑based \gls{VRDU} systems~\cite{Bai.2025,Luo.2024,Ye.2023} are typically built upon multimodal architectures that integrate image and text information in a dual‑encoder design. According to published model descriptions, these systems commonly employ a visual encoder (e.g., ViT‑ or Swin‑based backbones) alongside a textual encoder that processes externally derived OCR text. The resulting visual and textual representations are aligned through cross‑modal projection or fusion layers and subsequently passed to an LLM serving as the central reasoning module. This architectural pattern is extensively described in recent survey work~\cite{Ding.2025}.

Due to the substantially larger-scale pretraining of LLMs compared to Transformer‑based \gls{VRDU} models, these architectures are reported to retain extensive world knowledge, enabling improved generalization to diverse and structurally complex \glspl{VRD}~\cite{Rombach.2025}. A recent survey~\cite{Ding.2025} highlights that the rapid growth of LLM‑driven approaches in \gls{VRDU} is closely linked to their strong zero‑shot and few‑shot performance, resulting from broad pretraining and instruction tuning. These capabilities make LLMs appealing for processing complex \glspl{VRD} with minimal task‑specific supervision.

Nevertheless, a considerable performance gap in favor of fine‑tuned Transformer models is reported when compared to LLM‑based multimodal models, particularly on domain‑specific \gls{VRDU} tasks~\cite{Ding.2025,SciusBertrand.2024,Bhattacharyya.2025}. While LLMs offer flexibility and reduced reliance on fine‑tuning, they often fall short of specialized multimodal Transformers in tasks requiring high precision and robust layout understanding.

\subsection{OCR Integration}
In the context of scanned or photographed VRDs, document understanding systems operate on representations that must accurately expose the document’s textual content for downstream processing.
A commonly adopted strategy is to extract the text beforehand using an external OCR tool and provide it in digitized form to the classification model~\cite{Appalaraju.2024}. According to Scius‑Bertrand et al.~\cite{SciusBertrand.2024}, two principal methodological paradigms can be distinguished:
\begin{itemize}
	\item The first category comprises end-to-end (“OCR-free”) approaches that operate directly on document images and infer textual and structural cues from visual input representations, without relying on an explicit OCR stage. Its underlying assumption is that models can recover text semantics from visual patterns, thereby avoiding reliance on external \gls{OCR} frameworks.
	\item In contrast, the second category refers to \gls{OCR}‑dependent approaches, which integrate an external OCR framework to extract the document text prior to further processing. These systems detect text regions, segment them into units such as words or lines, and convert them into machine‑readable text together with associated layout metadata (e.g. bounding‑box coordinates). The resulting textual and positional features can then be used alone or jointly with the original image as visual representation for the model input.
\end{itemize}
Several studies~\cite{Ding.2025,SciusBertrand.2024} and models~\cite{Huang.2022,Appalaraju.2024,Luo.2024} report that incorporating OCR‑derived textual features can significantly improve performance on VRDs. Consistent with this observation, Rombach and Fettke~\cite{Rombach.2025} note that more than half of the LLM‑based document understanding approaches surveyed in their study integrate an external OCR engine, underscoring its continued relevance despite ongoing advances in OCR‑free document processing.
According to Jääskeläinen et al.~\cite{Jaaskelainen.2023}, commonly used OCR engines include Amazon Textract and open‑source alternatives such as Tesseract OCR, EasyOCR and PeroOCR.

\subsection{State‑of‑the‑Art Model Architectures}
The distinction between transformer-based and LLM-based multimodal architectures, with and without OCR integration, provides a conceptual overview of state-of-the-art architectures for document type classification and serves as the architectural reference framework for the methodology developed in Chapter~\ref{subsec:methodology}.

\subsubsection{Transformer-Based, OCR-Free}
These models operate directly on the document images. A visual encoder extracts visual feature representations, which are subsequently processed by Transformer layers~\cite{Rombach.2025}. 
Textual semantics is inferred implicitly from these visual representations. Effective use of such models typically requires task- and domain-specific fine-tuning.

\subsubsection{Transformer-Based, OCR-Dependent} 
In contrast, this category incorporates external \gls{OCR} outputs, like digitized text and optionally associated layout information, as additional input features.
The textual streams can be processed either (i) by a dedicated textual or layout-aware encoder or (ii) through token embeddings injected directly into the multimodal Transformer without a separate textual encoder~\cite{Ding.2025}. Visual encoders are commonly used in conjunction with the textual stream~\cite{Rombach.2025}.
As with the OCR-free counterparts, task- and domain-specific fine-tuning is typically required.

\subsubsection{LLM-Based, OCR-Free}
These models employ a visual encoder to extract feature representations from the document image, which are transformed into \gls{LLM}-compatible token embeddings through an adapter module. These visual tokens are fed into the \gls{LLM} as part of its input sequence, allowing its internal attention mechanisms to condition on visual evidence~\cite{Ding.2025}.
Task specification is performed via prompting (typically without task-specific fine-tuning), leveraging the \glspl{LLM} generalization and prior knowledge~\cite{Ding.2025,Rombach.2025}.

\subsubsection{LLM-Based, OCR-Dependent}
These models operate on OCR-extracted text (and optionally layout metadata) as their primary input. Integration of the textual signal typically follows one of two strategies: (i) OCR-derived text tokens are appended directly to the task prompt and processed through the \glspl{LLM} native tokenizer, or (ii) the OCR tokens are converted into embeddings and injected into the \gls{LLM} through an adapter module~\cite{Ding.2025,Rombach.2025,Appalaraju.2024}.
A visual encoder is often retained to provide complementary visual evidence~\cite{Rombach.2025}. Task specification is usually achieved through prompting instead of task-specific fine-tuning~\cite{Ding.2025,Rombach.2025}.

\subsection{Datasets for Type Classification}
RVL-CDIP (Ryerson Vision Lab Complex Document Information Processing)~\cite{Harley.2015} is a commonly used dataset for document type classification~\cite{Rombach.2025,Larson.2023,SouhailBakkali.2023,Skalicky.2022}. It comprises 400,000 greyscale document images, mainly derived from scanned documents, covering 16 document categories such as form, email, invoice, and specification. Each document is annotated with a single document-type label. The dataset is class-balanced, with 25,000 samples per category, and follows a predefined train/validation/test split of 320k/40k/40k. 

The IIT-CDIP dataset~\cite{Lewis.2006} comprises approximately six million document images and is derived from the Legacy Tobacco Document Library (LDL), a publicly released collection of internal documents from U.S. tobacco companies, including correspondence, reports, memoranda, and legal materials. RVL-CDIP constitutes a labeled subset of this dataset.

DocLayNet~\cite{Pfitzmann.2022} is a human-annotated dataset primarily designed for document layout segmentation, containing 80,863 document pages from heterogeneous sources. In addition to layout annotations, it provides document-level category labels for six classes (financial reports, scientific articles, laws and regulations, government tenders, manuals, patents). As a result, it can also be used for document type classification, although this task is not its primary focus.\medskip

In contrast, many datasets in the document understanding domain focus on single document categories and are primarily tailored to information extraction tasks rather than document type classification~\cite{Abdallah.2024}. As a result, they are generally unsuitable as standalone benchmarking datasets for this task. Representative examples include:
\begin{itemize}
	\item CORD~\cite{SeunghyunPark.2019}: Approximately 1,000 photographed receipts, mainly from supermarkets and restaurants, annotated for semantic entity recognition and relation extraction.
	\item FUNSD~\cite{Jaume.2019}: A dataset of 199 scanned English forms, annotated for key-value pair extraction.
	\item XFUND~\cite{Xu.2022}: A multilingual extension of FUNSD, consisting of 199 annotated forms each in Chinese, Japanese, Spanish, French, Italian, German, and Portuguese.
\end{itemize}

%% file: files/methodology.tex
\section{Methodology}\label{subsec:methodology}
This section introduces the methodological approach of the study. It outlines the analytical evaluation strategy, defines the architectural design space considered, and motivates the selection of one representative model for each architectural category.

\subsection{Task Definition and Study Design}
	The purpose of this study is to assess the ability of different multimodal Transformer- and \gls{LLM}-based architectures to interpret a document's content in a way that enables reliable type prediction. Document Type Classification is a single-label classification problem in which each \gls{VRD} must be assigned to exactly one predefined document category.
	
	Modern approaches can be categorized along two independent architectural dimensions that define the design space considered in this study: 
	\begin{enumerate}[(a)]
		\item the model backbone type, distinguishing Transformer‑based from \gls{LLM}-based architectures, and
		\item the feature integration strategy, differentiating \gls{OCR}-dependent from \gls{OCR}-free models.
	\end{enumerate}
	This two-dimensional taxonomy builds on prior architectural analyses~\cite{Ding.2025,SciusBertrand.2024} and reflects the dominant design choices underlying contemporary multimodal \gls{VRDU} systems.

	To enable a controlled comparison across this design space, we select one representative state-of-the-art model for each architectural category. The selected models are evaluated using RVL-CDIP~\cite{Harley.2015} as a publicly recognized and openly accessible curated benchmarking dataset for document type classification.

	This evaluation aims to:
	\begin{enumerate}
		\item quantify type classification performance with respect to accuracy and computational efficiency;
		\item systematically characterize architectural and implementation-level strengths and limitations across different model designs; 
		\item analyze the impact of \gls{OCR}-derived textual features on the performance of Transformer- and LLM-based models.
	\end{enumerate}
	All models are evaluated under identical experimental conditions, including uniform evaluation scripts, shared dataset splits, consistent prompting strategies for \gls{LLM}-centric architectures, and comparable inference configurations.

\subsection{Representative Model Selection}
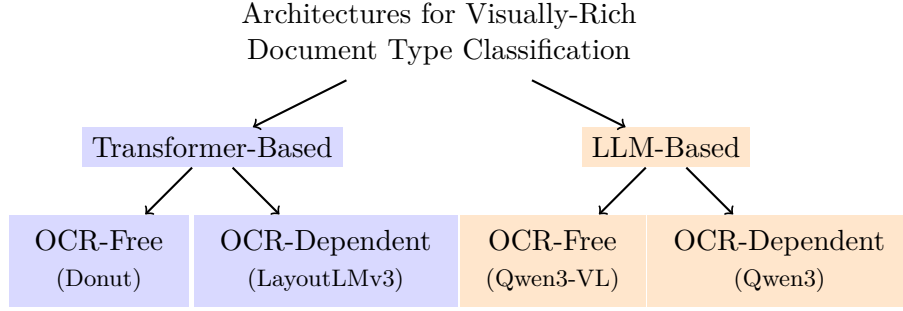
\begin{figure}[htbp]
	\centering
	\begin{tikzpicture}[
		node/.style={
			draw,
			rectangle,
			rounded corners,
			minimum width=3cm,
			minimum height=1cm,
			align=center,
			font=\small
		},
		trans/.style={fill=blue!15},
		llm/.style={fill=orange!20},
		arrow/.style={->, thick}
		]
		
		\node (root) at (0,0) {\begin{tabular}{c}
				Architectures for Visually-Rich\\
				Document Type Classification
			\end{tabular}	
		};
		
		\node[trans] (trans) at (-3,-1.5) {Transformer-Based};
		\node[llm]   (llm)   at ( 3,-1.5) {LLM-Based};
		
		\node[trans] (t_of) at (-4.5,-3){
			\begin{tabular}{c}
				OCR-Free \\ \footnotesize (Donut)
			\end{tabular}
		};
		
		\node[trans] (t_od) at (-1.5,-3){
			\begin{tabular}{c}
				OCR-Dependent \\
				\footnotesize (LayoutLMv3)
			\end{tabular}
		};
		
		\node[llm] (l_of) at ( 1.5,-3){
			\begin{tabular}{c}
				OCR-Free \\ \footnotesize (Qwen3-VL)
			\end{tabular}
		};
		
		\node[llm] (l_od) at ( 4.5,-3){
			\begin{tabular}{c}
				OCR-Dependent \\
				\footnotesize (Qwen3)
			\end{tabular}
		};
		
		\draw[arrow] (root) -- (trans);
		\draw[arrow] (root) -- (llm);
		
		\draw[arrow] (trans) -- (t_of);
		\draw[arrow] (trans) -- (t_od);
		
		\draw[arrow] (llm) -- (l_of);
		\draw[arrow] (llm) -- (l_od);
		
	\end{tikzpicture}
	\caption{Taxonomy of architectures for Visually-Rich Document Type Classification based on model type (transformer vs. LLM-based) and OCR integration (OCR-free vs. OCR-dependent).}
	\label{fig:model_taxonomy}
\end{figure}

The following models were selected as representative examples of the main architectural families in multimodal document understanding, as illustrated in Figure~\ref{fig:model_taxonomy}. Each model reflects a specific combination of backbone design and modality integration strategy, allowing a systematic comparison across these dimensions.

\subsubsection{Transformer-Based, OCR-Free}
Donut (Document Understanding Transformer)~\cite{Kim.2021} is an OCR-free Transformer model for \gls{VRDU} with openly available weights and fine-tuned variants on established benchmarking datasets, including RVL-CDIP for document type classification and CORD for information extraction. Reported results indicate competitive accuracy and inference time~\cite{Kim.2021}. 
Its availability as a pretrained foundation model under an open-source MIT license, along with its frequent use as a baseline in subsequent \gls{VRDU} studies~\cite{SciusBertrand.2024,Larson.2023,Appalaraju.2024}, positions it as a robust and widely adopted model in this category.

Instead of relying on external \gls{OCR} frameworks for text extraction, Donut employs solely a slightly modified SWIN Transformer~\cite{Liu.2021} as its visual encoder for input processing, combined with a multilingual BART~\cite{Lewis.2020,Liu.2020} model serving as the textual decoder. This decoder is pretrained using a teacher‑forcing scheme with a cross‑entropy loss objective.

\subsubsection{Transformer-Based, OCR-Dependent} 
LayoutLMv3~\cite{Huang.2022} is a general-purpose pretrained multimodal Transformer encoder designed for both text- and image-centric Document AI tasks.
Its architecture integrates three input modalities:
(i) text embeddings from \gls{OCR}‑derived word-token sequences using a RoBERTa‑initialized~\cite{Liu.2019} embedding layer,
(ii) layout embeddings encoding the 2D bounding‑box positions of these tokens, and
(iii) visual embeddings produced by a DiT‑initialized~\cite{Li.2022} image tokenizer that converts the document image into patch‑level tokens.
All modalities are processed jointly within a multimodal Transformer encoder.

LayoutLMv3 has become an established baseline in contemporary VRDU research, with numerous studies~\cite{Gbada.2025,Larson.2023,Luo.2024,Abdallah.2024} benchmarking this model against competitors on datasets such as FUNSD~\cite{Jaume.2019} (form understanding) and CORD~\cite{SeunghyunPark.2019} (key information extraction). 
The model achieves state-of-the-art or consistently strong performance across these tasks. It further combines a modular design with flexible integration of \gls{OCR} tokens and is available as an open-source implementation under the CC-BY-NC-SA-4.0 license.

\subsubsection{LLM-Based, OCR-Free}
Qwen3-VL~\cite{Bai.2025} is a modern \gls{LLM} designed for \gls{OCR}-free vision-language processing and long-context multimodal comprehension, making it suitable for \gls{VRDU} tasks. For this study, the dense 32B variant Qwen3-VL-32B-Instruct~\cite{Bai.2025} under Apache 2.0 license is used. It is the largest non-MoE model in the Qwen3-VL series and features 32 billion activated parameters per token.

Architecturally, the model consists of three components: (1) a SigLIP-2-based Vision Transformer~\cite{Tschannen.2025} serving as the vision encoder, (2) a two-layer \gls{MLP} as vision-language merger that compresses 2×2 encoder features into single visual tokens and includes additional specialized mergers for DeepStack multi-level decoder token injection, and (3) a dense Qwen3 language model decoder~\cite{Yang.2025} equipped with Interleaved-MRoPE for geometrically consistent multimodal positional encoding.
The training pipeline combines joint multimodal pretraining, integrating visual and textual knowledge on high-quality, curated data, with targeted post-training for instruction following and preference alignment, resulting in strong image-based reasoning performance.

\subsubsection{LLM-Based, OCR-Dependent}
Qwen3~\cite{Yang.2025} represents the text‑only LLM backbone of the Qwen3 family and is trained primarily for code generation, mathematical reasoning, and agentic tasks. In this study, we use the dense 32B variant Qwen3‑32B, released under the Apache‑2.0 license. Within the Qwen3 series, it is the largest non‑MoE model and, as with the Qwen3‑VL counterpart, activates 32 billion parameters per token.
Architecturally, Qwen3‑32B is a decoder‑only Transformer model equipped with self-attention with causal masks  and feed-forward neural networks (FFNs)~\cite{Yang.2024}.
It is a Transformer model with 64 layers, 64 query heads and 8 key/value heads, and supports a context length of 128K tokens. Unlike the multimodal Qwen3‑VL model, Qwen3‑32B does not include a vision encoder but processes text exclusively, using a byte‑level byte pair encoding tokenizer~\cite{Yang.2025}.

%% file: files/experimental_setup.tex
\section{Experimental Setup} \label{sec:experimental_setup}
All experimental results reported in this study are obtained under uniform and controlled evaluation conditions. Consistent hardware and software configurations, unified evaluation metrics, and identical data processing and inference protocols are applied to ensure comparability across architectures. These conditions are specified in detail in the following sections.

\subsection{Dataset}
	All experiments are conducted on the publicly available RVL-CDIP dataset~\cite{Harley.2015}, which has established itself as one of the most widely used benchmarks for document image classification and retrieval in \glspl{VRD}~\cite{Rombach.2025,Larson.2023,Skalicky.2022}.
	RVL-CDIP is particularly well suited for document type classification among publicly available datasets, as it covers a broad range of complex document types commonly encountered in business and administrative settings. The dataset comprises 16 coarse-grained document categories, including letter, memo, email, file folder, form, handwritten, invoice, advertisement, budget, news article, presentation, scientific publication, questionnaire, resume, scientific report, and specification. In contrast to datasets such as DocLayNet~\cite{Pfitzmann.2022}, which focus on layout understanding, RVL-CDIP provides higher-level document type labels that are directly aligned with the classification objective considered in this work.
	
	RVL-CDIP consists of 400{,}000 scanned greyscale document images, each annotated with a single document type label. The dataset follows a predefined training/validation/test split of 320k/40k/40k, with 25{,}000 samples per class, resulting in a balanced label distribution. All images are downscaled such that the length of the longest image side does not exceed 1000 pixels.
	
	For empirical evaluation, we use the entire test set of 40{,}000 document images, which exhibits an approximately uniform class distribution. The dataset is accessed via the HuggingFace \emph{datasets} repository\footnote{\url{https://huggingface.co/datasets/chainyo/rvl-cdip}}. All documents are provided as single-channel greyscale images using the dataset’s native \emph{PIL.Image} data structure.

\subsection{Evaluation Metrics}
	We evaluate all models using standard classification and runtime metrics: 
	\begin{itemize}
		\item \textbf{Accuracy:} A global measure of overall classification performance across all classes.
		\item \textbf{Classification report:} Precision, recall, and F1-scores for each label, capturing class-specific performance characteristics.
		\item \textbf{Confusion matrix:} A class-wise visualization of error patterns.
		\item \textbf{End-to-end inference time:} The total time required to transform all document images into their final predicted labels. This metric reflects the full inference pipeline, including OCR processing (for OCR-dependent models), image and text encoding, and the model’s forward pass.
	\end{itemize}
	Inference latency is measured using single-image classification (batch size~1) to ensure strict cross-model comparability.

\subsection{Hardware Setup}
All experiments are executed on 4 x NVIDIA H200 (141 GB) GPUs, paired with a dual-socket Intel Xeon Platinum 8462Y+ processor, providing a total of 64 physical cores and 128 hardware threads. The system provides 2 TB of memory and runs on Ubuntu 24.04.03 LTS.

\subsection{Software Setup}
To ensure comparability and reproducibility, all models are initialized using a unified execution script. 
This includes shared logging utilities, a consistent virtual environment, and a unified pipeline across all experiments. Each run records the corresponding git commit hash to guarantee full traceability of the exact code state used during execution.

The software environment is based on Python~3.12.3 and PyTorch~2.7.1+cu126, providing compatibility with GPU-accelerated components of the pipeline. Distributed and device-agnostic execution is orchestrated using \texttt{accelerate}~1.12.0, applied where required by specific model architectures. All evaluated models are loaded and executed through the \texttt{transformers}~5.0.0 library, which offers architecture-agnostic tooling for model initialization, tokenization, and inference \cite{Wolf.2020}. Dataset management is handled via \texttt{datasets}~3.6.0, ensuring consistent data access and caching across runs. The dataset and all model weights are stored and loaded locally to avoid variability from remote I/O or network latency.

\paragraph{OCR Setup} For OCR-dependent models, we rely on Tesseract~OCR~5.3.4 together with the \texttt{tessdata-eng} language pack for English text recognition.

\subsection{Model-Specific Setup}
To ensure reproducibility while maintaining comparability across architectures, model-specific setup configurations are defined for each evaluated system, as differences in modality integration require distinct and explicitly documented implementations. Implementations rely on official model-provided classes wherever possible, with deviations from the original model configuration introduced only when required by the underlying multimodal design.

\subsubsection{Donut}
The experiments use the official Donut checkpoint fine-tuned on RVL-CDIP\footnote{\url{https://huggingface.co/naver-clova-ix/donut-base-finetuned-rvlcdip}}, released on HuggingFace by the original authors. 

For preprocessing, the \texttt{DonutProcessor} is employed, combining (i) the \texttt{DonutImageProcessor} for image preprocessing and (ii) an XLM-Roberta tokenizer (via \texttt{XLMRobertaTokenizer}) for prompt and target tokenization.
For each input document, the image processor applies the standard Donut preprocessing pipeline, including resizing, thumbnail generation, zero-padding to the model’s expected spatial canvas, rescaling, and normalization. In parallel, the task prompt is tokenized using Donut’s prompting scheme. The resulting tensorized \texttt{pixel\_values} (image representations) and \texttt{input\_ids} (text prompt) are then passed jointly to the model.
Since Donut formulates classification as sequence-to-sequence generation conditioned on an instructional prompt, the model’s autoregressive decoder produces a structured output sequence that encodes the predicted class (typically a JSON-like string). This output sequence is subsequently parsed to extract the final label.

\subsubsection{LayoutLMv3}
No official RVL-CDIP fine-tuning checkpoint is provided for LayoutLMv3 by the model developers (Microsoft), despite RVL-CDIP results being reported in the original publication~\cite{Huang.2022}. We therefore use a publicly available HuggingFace checkpoint\footnote{\url{https://huggingface.co/gordonlim/layoutlmv3-base-finetuned-rvlcdip}}, based on the official LayoutLMv3-Base model\footnote{\url{https://huggingface.co/microsoft/layoutlmv3-base}}. This checkpoint showed the strongest performance among publicly available variants in our preliminary screening (see Table~\ref{tab:rvl_screening}).
\begin{table}[ht]
	\centering
	\footnotesize
	\begin{tabular}{p{11cm} c}
		\textbf{RVL-CDIP Checkpoint for LayoutLMv3 model} & \textbf{Accuracy} \\
		\hline
		\url{https://huggingface.co/gordonlim/layoutlmv3-base-finetuned-rvlcdip} & 0.90 \\
		\url{https://huggingface.co/davidhajdu/fine-tuned-rvl-cdip} & 0.83 \\
		\url{https://huggingface.co/felixtran/layoutlmv3-rvl-cdip-small} & 0.65 \\		
	\end{tabular}
	\caption{Preliminary screening results of publicly available LayoutLMv3 checkpoints fine-tuned on RVL-CDIP, used to select the best-performing model.}
	\label{tab:rvl_screening}
\end{table}
The inference pipeline proceeds as follows: All input images are converted to RGB, as greyscale formats are not supported by the model. The images are processed by the \texttt{LayoutLMv3Processor}, which delegates internally to the \texttt{LayoutLMv3ImageProcessor} for OCR-based text and layout extraction via Tesseract OCR, followed by resizing, rescaling and normalization. The OCR-derived word-level tokens and their associated bounding boxes are subsequently passed to the \texttt{LayoutLMv3Tokenizer}, which converts them into model-compatible token embeddings using a \texttt{max\_length} padding strategy and longest-first truncation with \texttt{max\_length = 512}. The visual, textual and bounding box embeddings are subsequently fused into the model-ready representation. All components operate under their default configurations unless stated otherwise.

\subsubsection{Qwen3-VL-32B-Instruct}
This study employs the official Qwen3-VL-32B-Instruct model\footnote{\url{https://huggingface.co/Qwen/Qwen3-VL-32B-Instruct}} from the HuggingFace Hub, without any additional fine-tuning.

Following model initialization, each document image is processed by the \texttt{Qwen3VLProcessor}, which delegates to the \texttt{Qwen2VLImageProcessorFast} for image preprocessing. This includes aspect-preserving smart resizing (subject to minimum-size and patch-grid divisibility constraints), followed by rescaling and normalization, producing model-ready pixel values for Qwen3‑VL’s SigLIP‑2-based vision encoder.

In parallel, the task prompt is tokenized using the \texttt{Qwen2Tokenizer} within the same processor.
It applies Qwen3‑VL's multimodal chat template, inserts the appropriate vision placeholder tokens (such as \emph{vision-start}, \emph{image}, and \emph{vision-end} for the image position), and converts the final prompt string into input IDs. The prompt used in our experiments is shown below:
\begin{Verbatim}[breaklines=true, breakanywhere=true, fontsize=\footnotesize]
	'Classify the document into one of the following types (output only the class name): letter, form, email, handwritten, advertisement, scientific\_report, scientific\_publication, specification, file\_folder, news\_article, budget, invoice, presentation, questionnaire, resume, memo
	
	Document:
	"{document\_text}"'
\end{Verbatim}
The preprocessed pixel values and prompt token ids are jointly fed to the model. Generation proceeds with a maximum of 128 new tokens. Unless stated otherwise, all processor and generation settings follow their default configuration.

\subsubsection{Qwen3-32B}
This study uses the official Qwen3‑32B model\footnote{\url{https://huggingface.co/Qwen/Qwen3-32B}} from the HuggingFace Hub without any additional fine‑tuning.

Document text is externally obtained using Tesseract OCR via the \texttt{image\_to\_data} function from the \texttt{pytesseract} module. From the resulting OCR output, only word-level text is retained as a list. These items are concatenated into a single string and inserted into the task prompt at the position of the \texttt{\{ocr\_text\}} placeholder: 
\begin{Verbatim}[breaklines=true, breakanywhere=true, fontsize=\footnotesize]
	'The text below was extracted from a document image using OCR.
	
	OCR Text:
	"""
	{ocr_text}
	"""
	
	Classify the document into one of the following types:
	
	letter, form, email, handwritten, advertisement, scientific_report, scientific_publication, specification, 
	file_folder, news_article, budget, invoice, presentation, questionnaire, resume, memo
	
	Return exactly one label from the list, using lowercase ASCII letters only, with no spaces, no punctuation, no quotes.
	
	Output format (exactly this): 
	<label>
	'
\end{Verbatim}
This prompt is processed using the \texttt{Qwen2Tokenizer}, which applies the model’s chat template through its Jinja‑based rendering mechanism (\texttt{tokenize=False}, \texttt{add\_generation\_prompt=True}, \texttt{enable\_thinking=False}). 
After rendering, the final prompt is tokenized into input IDs and fed to Qwen3‑32B for generation.
All experiments use a maximum of 128 new tokens, and all remaining generation parameters follow their default settings unless stated otherwise.

%% file: files/results.tex
\section{Results} \label{sec:results}
This section presents the empirical results of our evaluation. 
An initial interpretation of the overall performance of all models on the RVL‑CDIP benchmark is followed by a model‑class‑specific analysis highlighting the characteristic strengths and weaknesses of the evaluated architectures. Finally, practical aspects related to model usability in applied settings are discussed.

During evaluation, one dataset entry (index~34965, test split) caused an exception (\texttt{Exception: cannot identify image file <\_io.BytesIO object>}). The corresponding image could not be processed and is therefore excluded from this study.

\subsection{Overall Evaluation Results}
The following subsection analyzes the overall performance of the models in terms of accuracy and inference time.
Table~\ref{tab:perclass_metrics_TC} summarizes the accuracy and the per-class precision, recall, and F1-scores obtained under identical inference conditions for all four evaluated models. 
Additionally, the end‑to‑end inference time is reported, measured over the full prediction pipeline, including \gls{OCR} for \gls{OCR}-dependent models.
\begin{sidewaystable}
	\centering
	\small
	\begin{tabular}{l|c|ccc|ccc|ccc|ccc}
		& 
		& \multicolumn{3}{c|}{\textbf{Donut}}
		& \multicolumn{3}{c|}{\textbf{LayoutLMv3}}
		& \multicolumn{3}{c|}{\textbf{Qwen3-VL-}}
		& \multicolumn{3}{c}{} \\
		
		& 
		& \multicolumn{3}{c|}{(RVL-CDIP \glsentryshort{FT})}
		& \multicolumn{3}{c|}{(RVL-CDIP \glsentryshort{FT})}
		& \multicolumn{3}{c|}{\textbf{32B-Instruct}}
		& \multicolumn{3}{c}{\textbf{Qwen3-32B}}\\
		
		\cline{3-5}
		\cline{6-8}
		\cline{9-11}
		\cline{12-14}
		
		\textbf{Class} & \textbf{\#Images}
		& \textbf{prec} & \textbf{rec} & \textbf{F1}
		& \textbf{prec} & \textbf{rec} & \textbf{F1}		
		& \textbf{prec} & \textbf{rec} & \textbf{F1}
		& \textbf{prec} & \textbf{rec} & \textbf{F1} \\

		\noalign{\hrule height 1.1pt}
		
		letter                & 2464 & 0.94 & 0.93 & 0.93 & 0.92 & 0.90 & 0.91 & 0.80 & 0.65 & 0.72  & 0.45 & 0.70 & 0.55 \\
		memo                  & 2492 & 0.96 & 0.96 & 0.96 & 0.94 & 0.91 & 0.93 & 0.43 & 0.96 & 0.59  & 0.31 & 0.82 & 0.45 \\ 
		email                 & 2516 & 0.99 & 0.99 & 0.99 & 0.98 & 0.99 & 0.99 & 0.99 & 0.82 & 0.90  & 0.99 & 0.61 & 0.75 \\
		file folder           & 2527 & 0.95 & 0.98 & 0.97 & 0.90 & 0.95 & 0.93 & 0.75 & 0.76 & 0.76  & 0.40 & 0.23 & 0.29 \\
		form                  & 2506 & 0.91 & 0.89 & 0.90 & 0.77 & 0.83 & 0.80 & 0.52 & 0.51 & 0.51  & 0.36 & 0.22 & 0.27 \\
		handwritten           & 2532 & 0.96 & 0.97 & 0.97 & 0.91 & 0.95 & 0.93 & 0.67 & 0.86 & 0.75  & 0.32 & 0.48 & 0.38 \\ 
		invoice               & 2477 & 0.96 & 0.96 & 0.96 & 0.91 & 0.81 & 0.86 & 0.79 & 0.78 & 0.79  & 0.59 & 0.57 & 0.58 \\ 
		advertisement         & 2515 & 0.95 & 0.96 & 0.96 & 0.89 & 0.91 & 0.90 & 0.91 & 0.87 & 0.89  & 0.79 & 0.32 & 0.46 \\
		budget                & 2505 & 0.95 & 0.97 & 0.96 & 0.89 & 0.79 & 0.84 & 0.83 & 0.61 & 0.70  & 0.72 & 0.36 & 0.48 \\ 
		news article          & 2463 & 0.96 & 0.95 & 0.95 & 0.82 & 0.92 & 0.87 & 0.73 & 0.88 & 0.80  & 0.69 & 0.67 & 0.68 \\ 
		presentation          & 2489 & 0.92 & 0.91 & 0.91 & 0.87 & 0.85 & 0.86 & 0.92 & 0.20 & 0.33  & 0.79 & 0.09 & 0.16 \\ 
		scientific publication& 2571 & 0.97 & 0.95 & 0.96 & 0.96 & 0.91 & 0.93 & 0.86 & 0.93 & 0.89  & 0.83 & 0.80 & 0.81 \\
		questionnaire         & 2435 & 0.95 & 0.94 & 0.95 & 0.82 & 0.89 & 0.86 & 0.98 & 0.74 & 0.85  & 0.97 & 0.66 & 0.78 \\ 
		resume                & 2537 & 0.99 & 0.99 & 0.99 & 0.99 & 0.98 & 0.99 & 1.00 & 0.97 & 0.98  & 0.98 & 0.79 & 0.88 \\
		scientific report     & 2498 & 0.90 & 0.92 & 0.91 & 0.85 & 0.84 & 0.85 & 0.72 & 0.58 & 0.64  & 0.47 & 0.63 & 0.54 \\
		specification         & 2472 & 0.97 & 0.97 & 0.97 & 0.97 & 0.90 & 0.93 & 0.80 & 0.87 & 0.83  & 0.58 & 0.82 & 0.68 \\
		
		\noalign{\hrule height 1.1pt}
		
		\textbf{Accuracy}   
		&  
		& \multicolumn{3}{c|}{0.95} 
		& \multicolumn{3}{c|}{0.90} 
		& \multicolumn{3}{c|}{0.75}  
		& \multicolumn{3}{c}{0.55} \\
		\noalign{\hrule height 1.1pt}
		\textbf{\shortstack[l]{Mean end-to-end\\inference time\\per image}}   
		&  
		& \multicolumn{3}{c|}{307 ms} 
		& \multicolumn{3}{c|}{554 ms} 
		& \multicolumn{3}{c|}{239 ms} 
		& \multicolumn{3}{c}{600 ms}\\
		
		\noalign{\hrule height 1.1pt}
	\end{tabular}
	\captionsetup{width=0.75\textheight}
	\caption{Per-class precision (prec), recall (rec), and F1-scores (f1) of the four evaluated models on the RVL-CDIP test split. Accuracy and average end-to-end inference time per model are reported below.}
	\label{tab:perclass_metrics_TC}
\end{sidewaystable}

\subsubsection{Accuracy}
Comparing the accuracy values reported by the model developers (LayoutLMv3: 0.9544~\cite{Huang.2022}, Donut: 0.9530~\cite{Kim.2021}) with our results obtained in this study indicates that Transformer models specifically optimized for RVL‑CDIP clearly outperform the evaluated LLMs (Qwen3‑VL: 0.75, Qwen3: 0.55). 

This comparison is complicated by the unavailability of the official LayoutLMv3 checkpoints. Consequently, this study relies on a publicly released fine‑tuned variant (accuracy: 0.90) rather than the original specialized model. The selected checkpoint was identified as the strongest among three publicly available fine‑tuned versions, while the remaining two performed substantially worse (see Table~\ref{tab:rvl_screening}).

The measured accuracy of 0.90 for this fine-tuned LayoutLMv3 checkpoint illustrates that fine-tuning is a non‑trivial process with substantial influence on model performance. Achieving accuracy levels reported in the original LayoutLMv3 paper requires near‑optimal fine‑tuning, which is computationally demanding and depends on both methodological expertise and sufficient resources. Without such conditions, Transformer models do not consistently surpass general‑purpose LLMs. 
It is also notable that the OCR‑free Transformer Donut achieves an accuracy of 0.95 in this evaluation, which matches the performance of the OCR‑dependent LayoutLMv3 variant reported in the literature (0.9544)~\cite{Huang.2022}.
On the LLM side, the reported results indicate that processing the document image directly, rather than relying solely on OCR‑extracted text, provides a measurable advantage for this classification task.

\subsubsection{End-to-end inference times}
The results show clear differences in end-to-end inference time. The vision-language LLM model Qwen3‑VL‑32B is the fastest, with an average of 239 ms per document image (9,545.20s total), followed closely by Transformer Donut at 307 ms (12,296.66 s total); both models are OCR-free. The OCR-dependent models perform noticeably slower: LayoutLMv3 requires in average 554 ms (22,144.26 s total) and Qwen3‑32B 600 ms per image (23,981.41 s total). It is important to note that these end‑to‑end times for LayoutLMv3 and Qwen3‑32B include the OCR stage. This raises the question of whether the substantially longer runtime of the OCR‑dependent models is primarily attributable to the OCR step itself.

\subsubsection{Unknown labels}
As an encoder‑only architecture, LayoutLMv3 does not perform classical text generation. Rather than producing free-form outputs, its multimodal representations are processed by a classification head that maps them to a predefined label set via a linear layer and softmax function. Consequently, the absence of a decoding stage inherently restricts the model’s outputs to this fixed set of document type labels.
Donut, in contrast, employs an autoregressive Transformer decoder that could in principle generate arbitrary output token sequences. 
In practice, the model reliably produces valid class labels as fine‑tuning on the predefined label set effectively constrains the decoder to the target label space.

For the LLM‑based models, we observe a different behavior than for the Transformer models. Since no fine‑tuning is applied, the task definition and the set of allowed labels were specified solely through the prompt. As a result, both LLMs generate labels outside the predefined target set (Qwen3‑VL: 37 invalid labels; Qwen3: 972 invalid labels), which reduce the recall. Inspection of the corresponding misclassified samples shows that many of these documents are difficult to assign unambiguously to a single class and that the predicted label often coincides with salient text appearing in the document itself. For example, image~8162 (RVL‑CDIP test split) belongs to the file\_folder class but contains the phrase “Database marketing proposal”, which leads the model to predict proposal. This raises the question of whether LLMs tend to rely more strongly on prominent textual fragments than on global document semantics.

\subsection{Model-Class Analysis}
The following subsection presents a class‑level analysis of the individual models.

\subsubsection{Donut}
Donut achieves its strongest performance in categories with clear visual and structural distinctiveness. In particular, the email, resume, specification, handwritten, and file\_folder classes reach F1-scores between 0.97 and 0.99. Manual inspection confirms that documents in these categories are visually well defined and can be identified with a high degree of certainty.

\begin{figure}
	\centering
	\includegraphics[width=0.8\textwidth]{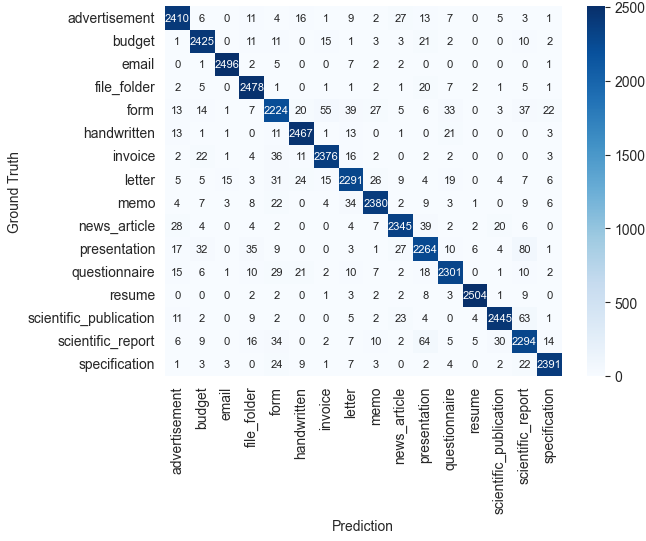}
	\caption{Donut: Confusion matrix on the RVL‑CDIP test split.} \label{conf_Donut_TC}
\end{figure}

Lower‑performing categories fall within an F1 range of 0.90–0.91, most notably form, presentation, and scientific\_report. The form class shows the clearest weakness, reflected in its recall of 0.89 and a wide dispersion of misclassifications. A manual inspection of this class reveals that many form documents are not visually well separated from other document type layouts and could plausibly fit into several categories.
This is consistent with the confusion matrix (Fig.~\ref{conf_Donut_TC}), where form documents are assigned to classes with similar semi-structured layouts such as invoice (55), letter (39), scientific\_report (37), or questionnaire (33).
A similar pattern is observed for the presentation class. Errors cluster toward scientific\_report (80) and file\_folder (35), which share generic layout characteristics. Presentation slides often contain title-page fronts, prominent headings, and block-structured content, which can resemble the visual organization of reports or folder-style documents.

\subsubsection{LayoutLMv3}
LayoutLMv3 also demonstrates strong classification performance. The model achieves F1-scores of 0.99 for both email and resume and consistently high scores for file\_folder (0.93), handwritten (0.93), memo (0.93), specification (0.93), and scientific\_publication (0.93).

\begin{figure}
    \centering
	\includegraphics[width=0.8\textwidth]{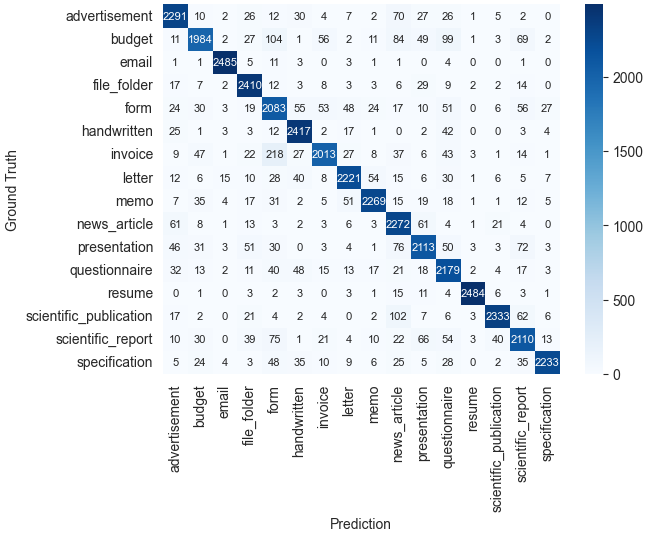}
	\caption{LayoutLMv3: Confusion matrix on the RVL‑CDIP test split.} \label{confusion_LayoutLMv3_TC}
\end{figure}

Both budget and invoice exhibit comparatively low recall but higher precision, indicating conservative label assignment and a tendency to miss true instances. The confusion matrix (Fig.~\ref{confusion_LayoutLMv3_TC}) further illustrates this pattern: budget documents are frequently predicted as invoice (47), memo (35), or form (30). Conversely, invoice documents are often classified as budget (56) or form (53). A plausible explanation is that many budget and invoice documents share highly similar template-like layouts. As a result, instances that deviate from these typical patterns may not be reliably recognized and are instead assigned to visually related categories.

In contrast, the news\_article and questionnaire classes exhibit the opposite pattern, characterized by lower precision but comparatively higher recall, indicating a tendency to attract a substantial number of false positives.
For the news\_article class, advertisement (61) and presentation (61) documents are frequently misclassified into this category, indicating that prominent headings or mixed text–image layouts may lead the model to interpret them as news content.
A similar effect appears for questionnaire  documents, with misclassifications from handwritten (48) and form (40), likely due to shared features such as handwritten elements or grid‑structured fields.

A pattern similar to that observed for Donut emerges for the presentation class, which again shows a broad dispersion across visually related classes. Presentation documents are therefore frequently assigned to news\_article (76) and advertisement (46), reflecting an overlap in headline‑driven layout characteristics. The form class exhibits a comparable behavior to that observed for Donut: documents in this category are not clearly separable from neighboring classes, resulting in lower precision (0.77) and recall (0.83).

\subsubsection{Qwen3-VL-32B-Instruct}
Qwen3‑VL attains strong F1‑scores on resume (0.98) and email (0.90), with solid results for advertisement (0.89) and scientific\_publication (0.89). 

As with the other models, the form class remains difficult to separate from neighboring labels. Its precision (0.52) and recall (0.51) indicate weak class boundaries and systematic spill-over into visually similar document types.

\begin{figure}
    \centering
	\includegraphics[width=0.8\textwidth]{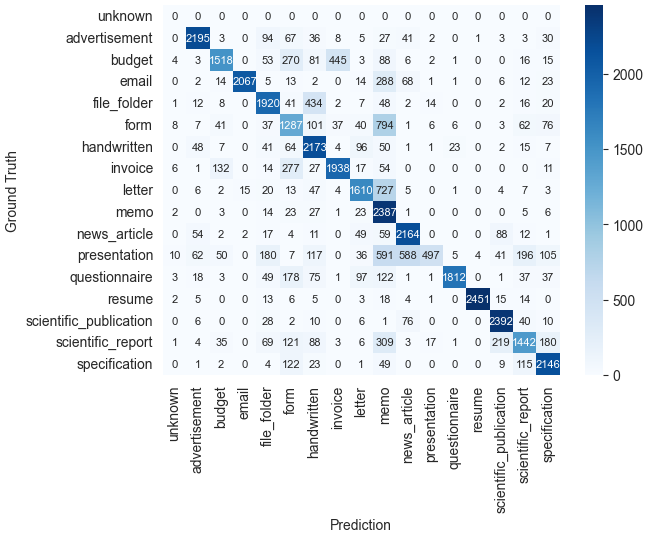}
	\caption{Qwen3‑VL‑32B‑Instruct: Confusion matrix on the RVL‑CDIP test split.} \label{conf_Qwen_VL_TC}
\end{figure}

The presentation class is markedly under‑predicted, with a precision of 0.92 but a substantially lower recall of 0.20. The model assigns this label only in high‑confidence cases, missing the majority of true presentation samples. 
As illustrated by the confusion matrix (Fig.~\ref{conf_Qwen_VL_TC}), many presentation documents are instead classified as memo (591) or news\_article (588), whereas only 497 instances are correctly identified. One possible explanation is that presentation-style layouts are underrepresented in the model’s training data, which may constrain its ability to reliably recognize this category.

In addition, Qwen3‑VL over-predicts the memo class. Documents from several categories, including form (794), letter (727), and presentation (591), are frequently misclassified as memo, resulting in a low precision of 0.43. At the same time, true memo samples are detected reliably, as reflected in a high recall of 0.96.

\subsubsection{Qwen3-32B}
\begin{figure}
    \centering
	\includegraphics[width=0.8\textwidth]{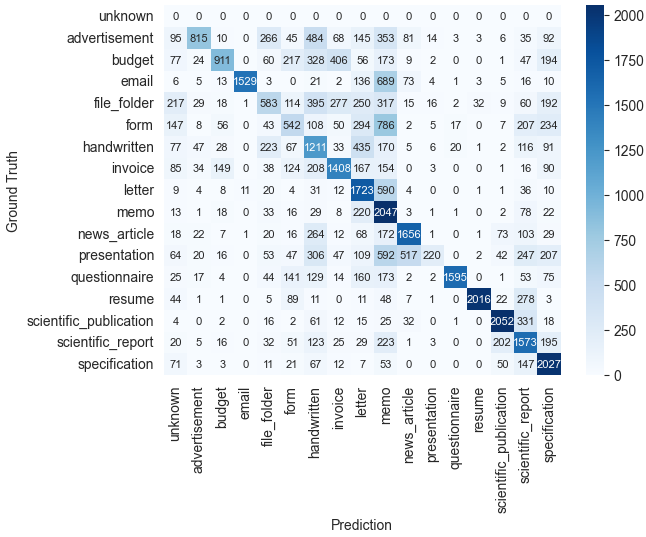}
	\caption{Qwen3-32B: Confusion matrix on the RVL‑CDIP test split.} \label{conf_Qwen_TC}
\end{figure}

Qwen3 achieves comparatively strong F1-scores in the resume (0.88) and scientific\_publication (0.81) classes. In contrast, the confusion matrix (Fig.~\ref{conf_Qwen_TC}) shows that several categories perform substantially worse. The file\_folder class (0.29) contains many documents with little text, while the form class (0.27) shows weak visual separability. Performance on handwritten documents (0.38) is reduced due to the loss of handwriting-specific cues during OCR extraction, and the presentation class (0.16) is primarily affected by its very low recall (0.09).

Recall is likewise low for advertisement (0.32) and form (0.22), whereas categories such as scientific\_publication (0.80) and memo (0.82) achieve higher recall, likely due to their more continuous text that the model can leverage. Precision is high for email (0.99), questionnaire (0.97), and resume (0.98), all of which exhibit distinctive lexical patterns. In contrast, precision is low for memo (0.31) and handwritten (0.32), where such stable textual cues are largely absent.

\subsection{Practical Usability}
In this study, all Transformer- and LLM‑based models could be integrated through Python’s \texttt{transformers} library. This resulted in a largely uniform workflow for model loading, input preprocessing such as tokenization, and inference, and was therefore straightforward to use from a user perspective. This kept the implementation effort manageable and ensured methodological consistency across models.
Occasional technical issues, for example related to external dependencies such as the \texttt{detectron2} module used in the Donut model, did occur but were resolved using standard installation and compatibility workarounds and did not impede the practical use of the models.

OCR‑dependent models required a more complex pipeline, as \texttt{Tesseract} as external OCR tool had to be integrated before inference. This introduced additional processing steps, dependencies, and maintenance overhead, making their practical use more involved than OCR‑free approaches.

We also observed substantial differences in resource demands. The LLM‑based models required considerably more memory and compute capacity than the Transformer‑based models, which limits their usability on standard hardware and makes experimentation more constrained. 

Another practical distinction concerns task specification. Transformer models rely on \linebreak fine-tuning, which may not always be available for domain‑ or language‑specific applications, while LLM‑based models normally depend entirely on prompting. In practice, the output quality was highly sensitive to the exact prompt formulation, and prompts became increasingly difficult to design when OCR‑extracted text had to be incorporated. This sensitivity made the effective use of LLM‑based models more demanding than the comparatively straightforward workflows of the Transformer‑based approaches.

%% file: files/discussion.tex
\section{Discussion} \label{sec:discussion}
The aim of this study was to examine how modern multimodal Transformer- and LLM-based architectures perform on document type classification under controlled conditions. This discussion relates the findings to the study’s objectives by addressing overall performance differences, architectural characteristics, and the broader implications for VRDU research and applied document type analysis.

\subsection{Multimodality for layout understanding}
Across all evaluated architectures, the results show that multimodal Transformer models continue to outperform LLM‑based approaches in document type classification. This holds in particular when considering the accuracy reported for the official LayoutLMv3 model, even though the corresponding checkpoint was not publicly available and therefore not reproducible in the present study. Notably, the publicly available LayoutLMv3 checkpoint evaluated here (accuracy: 0.90) falls noticeably short of the expected accuracy reported in the literature (0.9544), indicating a substantial performance gap when high-quality fine-tuning is not available.
As a result, the direct comparison between OCR-free and OCR-dependent Transformer variants should be interpreted with caution. 

The comparison between the vision‑language LLM Qwen3‑VL (OCR‑free) and the text‑only Qwen3 model (OCR‑dependent) shows that processing the document image directly is markedly superior to relying solely on OCR‑extracted text. This result underscores the importance of approaches that explicitly incorporate document layout, particularly for document types with strong layout‑dependent characteristics.

\subsubsection{Statement 1: Multimodality in linear document types.}
The results indicate that multimodality is not equally important across all document categories. Document types with predominantly linear structure and distinctive textual patterns tend to rely less on multimodal processing, as their content is conveyed largely through continuous and unambiguous text. 
This effect is particularly evident in the email category, where all four models achieve strong performance. The Transformer‑based architectures reach near‑perfect F1-scores (Donut and LayoutLMv3: 0.99), while the LLM‑based models maintain very high precision (Qwen3‑VL and Qwen3: 0.99) with only moderately reduced recall (Qwen3‑VL: 0.82, Qwen3: 0.61). Although the confusion matrices show misclassifications into textually similar classes such as memo (Qwen3‑VL: 288, Qwen3: 689), overall performance remains robust. This indicates that linear, text‑dominant document types rely less on multimodal cues and can be effectively classified based on textual features alone.

\subsubsection{Statement 2: Keyword-dependent performance.}
The results indicate that strongly structured, table-like document types can be classified reliably when they contain characteristic keywords or clearly identifiable textual patterns. This behavior is particularly evident in the resume category. Both Transformer-based models and the vision–language LLM Qwen3‑VL achieve strong performance on this class (LayoutLMv3 and Donut: F1 = 0.99; Qwen3‑VL: F1 = 0.98), and even the text-only LLM Qwen3 attains very high precision (0.98) despite the inherently layout‑encoded structure of resumes. An inspection of the document images suggests that resumes often include characteristic keywords that support correct identification. However, when such textual indicators are less prominent or occur in inconsistent form, the text-only model fails to detect the resume class reliably. This results in reduced recall (0.79) and misclassifications into related categories such as scientific\_report (278 cases) and form (89 cases).

\subsubsection{Statement 3: Well-defined and well‑separated label sets.}
The results demonstrate that classification performance strongly depends on how well a document category is defined and separable from neighboring classes. Categories with vague boundaries or heterogeneous layouts are consistently more difficult to identify across all evaluated models. This is evident, for example, in the form category, which shows below‑average precision and recall for every model (Donut: F1 = 0.90, LayoutLMv3: F1 = 0.80, Qwen3‑VL: F1 = 0.51, Qwen3: F1 = 0.27). The confusion matrices indicate that form documents overlap both in content and layout structure with several related classes, leading to frequent misclassifications. These findings highlight that reliable classification requires a well-defined and well‑separated label set, as broad or visually inconsistent categories reduce performance across architectures.

\subsubsection{Statement 4: Heterogeneous layouts.}
The results show that document types with heterogeneous or weakly standardized layouts are difficult to classify reliably, as they provide neither consistent layout patterns nor distinctive textual cues that models can exploit.
This effect is exemplified by the presentation category. Both Transformer-based models achieve only moderate precision and recall (Donut: precision 0.92, recall 0.91; LayoutLMv3: precision 0.87, recall 0.85) and frequently misclassify this category as visually related types such as news\_article. The LLM-based models show an even more degradation in performance, characterized by very low recall (Qwen3‑VL: 0.20, Qwen3: 0.09) and almost no correct assignments.
These observations suggest that the layout characteristics typical of presentation slides are not sufficiently learned by the models and that the available textual content alone does not provide a strong enough signal for accurate classification.
A plausible explanation is the limited presence of slide‑style documents in the filtered, high‑quality training data described in the Qwen3 technical reports~\cite{Bai.2025,Yang.2025}, which may have reduced the models’ exposure to this category.

\subsubsection{Statement 5: Image representation.}
The results show that multimodality, and in particular the use of an explicit image representation, is essential for document types in which class‑defining information is carried primarily by layout rather than text. Categories with little continuous text, few distinctive textual markers, or content that cannot be reliably captured by OCR therefore pose challenges for text‑only models such as Qwen3.
This effect is evident in the file\_folder and handwritten classes. In these cases, the text‑only LLM Qwen3 either lacks sufficient textual cues or cannot access the visual structure that characterizes handwritten content. In contrast, multimodal models that process the document image directly achieve substantially higher performance. 
These observations confirm that effective document type classification for layout‑driven documents depends on image‑based representations rather than solely on OCR‑extracted text.

\subsection{Model task adaptation strategies (fine-tuning vs. prompting)}
A notable observation is the substantial gap between the accuracy reported for the original LayoutLMv3 model~\cite{Huang.2022} and the performance achieved by publicly available fine‑tuned checkpoints, including the variant used in this study.
None of these checkpoints approaches the reported value of 95.44\%, indicating that fine‑tuning multimodal Transformer models is highly sensitive to training conditions and difficult to reproduce in practice. This strong dependence on high‑quality fine‑tuning represents a clear limitation in applied settings.

In contrast, LLM‑based models do not require fine‑tuning and can be used directly via prompting, which simplifies deployment across domains. This advantage comes at the cost of reduced control: prompt‑based interaction introduces variability in model outputs, reducing consistency and increasing the likelihood of predictions that do not conform to the predefined label set.

Overall, Transformer models provide reliable performance when well‑tuned, while LLMs offer ease of use but are less reliable in closed‑set classification. This trade‑off reflects the fundamental difference between the two approaches. Transformers benefit from task‑specific optimization but rely on high‑quality fine‑tuning, whereas LLMs can be applied immediately but lack the robustness and predictability required for strict label‑controlled classification tasks.

\subsection{Inference efficiency and the cost of OCR integration}
OCR-dependent pipelines introduce additional runtime overhead due to the OCR stage, even when the underlying Transformer or LLM backbone is computationally efficient. This reduces their suitability for large-scale or latency-sensitive document processing. In addition, OCR-induced errors propagate into the downstream model and may be amplified during prediction, further limiting robustness. OCR-free models avoid this bottleneck and achieve more consistent end-to-end inference times. 
The competitive runtime of the OCR-free Transformer Donut demonstrates that such architectures can achieve inference efficiency comparable to OCR-dependent models such as LayoutLMv3.

Although the OCR-only LLM Qwen3 shows that an extraction based solely on OCR suppresses layout, typography, and handwriting-specific cues and therefore performs poorly on visually driven classes such as handwritten documents, combining OCR text with the input image can still improve accuracy. This holds in particular when both modalities are processed jointly, as in the LayoutLMv3 model, allowing the model to exploit complementary visual and textual information.

\subsection{Dataset considerations}
Several RVL‑CDIP classes show substantial intra-class variability in layout and structure, particularly in categories such as form and questionnaire.
Certain categories exhibit substantial inter-class similarity, such as the visual overlap between presentation and news\_article or between invoice and budget documents. 

Similar limitations of the RVL‑CDIP dataset have been reported by Larson et al.~\cite{Larson.2023}, who identify considerable label noise (8.1\%), the presence of ambiguous or multi-label documents (1.7\%), and significant overlap between training and test splits, with at least 32\% of test samples having a duplicate or template-matched counterpart in the training set. The absence of published labeling guidelines further introduces additional uncertainty into the dataset.

Rombach et al.~\cite{Rombach.2025} highlight limitations of the dataset, noting that many documents originate from the 1990s and are characterized by poor image quality, scanning artefacts, and generally low resolution. These factors further limit the suitability of RVL‑CDIP as a reliable benchmark.

%% file: files/conclusion.tex
\section{Conclusion} \label{sec:conclusion}

\paragraph{Study Overview}
This study provides a systematic analysis of multimodal approaches to type classification on visually rich documents. It combines a structured review of common multimodal design strategies with an empirical comparison of two major paradigm families: specialized Transformer‑based models and general‑purpose LLM‑based models. 
The central objective was to understand how different multimodal input features such as text, images and layout information affect classification performance, with particular attention to the contrasting design principles of OCR‑dependent and OCR‑free architectures. Across all evaluated models, attention mechanisms play the central role in combining these modalities into a unified representation, enabling the integration within a single processing pipeline. To this end, four representative models were evaluated under a harmonized experimental pipeline on the RVL‑CDIP benchmark: LayoutLMv3 \textit{(OCR‑dependent Transformer)}, Donut \textit{(OCR‑free Transformer)}, Qwen3‑VL‑32B‑Instruct \textit{(OCR‑free LLM)} and Qwen3‑32B \textit{(OCR‑dependent LLM)}.

\paragraph{Key Findings}
The evaluation shows that well‑tuned Transformer models consistently outperform the LLM‑based approaches in document type classification on visually rich documents. Although both model families rely on attention mechanisms, their effectiveness depends strongly on the specific input modalities they receive.

Overall, the results indicate that visual features are the key driver for robust type classification, while OCR‑derived text serves mainly as complementary information. OCR‑free models such as Donut and Qwen3‑VL, which operate directly on the document image, achieve substantially reliable results across diverse document types. 
At the same time, OCR-dependent pipelines such as LayoutLMv3 benefit from the integration of explicit textual and positional features, particularly in text-rich or semi-structured documents. This advantage comes at the cost of increased inference time, additional implementation effort, and potential error propagation from imperfect OCR text extraction. 
Models that rely exclusively on OCR text, such as Qwen3‑32B, achieve acceptable performance only on documents with linear text flow but fail on typical VRDs with more demanding layout structure, such as documents containing tables, handwriting or layout‑encoded information, as they lack access to visual or spatial features. 

Specialized Transformer models provide the highest classification performance but require domain-specific fine-tuning. This process is computationally demanding and requires substantial practitioner expertise, as the base models must be adapted to the specific characteristics of the target domain, including document type, language, industry context, and task requirements. Publicly available LayoutLMv3 checkpoints further illustrate that insufficient fine-tuning can significantly degrade performance.

LLM‑based systems, while offering flexibility and rapid adaptation through prompting, remain sensitive to variations in prompt formulation, are more challenging to deploy in privacy‑constrained on‑premise environments due to their computational requirements, and generally deliver weaker performance on layout‑heavy \glspl{VRD}. Observations further suggest that LLMs tend to prioritize salient textual fragments rather than fully exploiting underlying layout structure.

Finally, the study underscores the importance of carefully curated datasets with document classes that are clearly separated in both visual structure and content. 
The findings indicate that when document types overlap or vary strongly in their internal structure, models struggle to learn reliable class-defining features. Effective training therefore requires document categories that are both consistent within each class and clearly distinguishable from one another.

\paragraph{Contributions}
This study closes a central gap in VRDU research by providing a consistent and transparent evaluation of multimodal state‑of‑the‑art models for document type classification. Unlike prior work that primarily reports aggregate accuracy on RVL‑CDIP, the analysis offers a fine‑grained comparison across individual classes and layout characteristics, enabling a clearer understanding of modality‑specific strengths and limitations.

The findings show that specialized multimodal Transformers remain superior to general‑purpose LLMs on visually rich and layout‑intensive documents, and that visual processing is essential for reliable type classification. 
These results provide a more systematic basis for evaluating multimodal design choices and support informed model selection in VRDU.

\paragraph{Implications and Outlook}
From a practical perspective, no single architecture can be recommended universally. The choice depends on the required classification accuracy, deployment constraints and existing expertise. Specialized multimodal Transformer models currently deliver the strongest results but require domain‑specific fine‑tuning and substantial computational resources. 
LLM‑based systems enable rapid adaptation through prompting but remain less reliable on structured VRDs and are more difficult to operate on‑premise due to their high hardware requirements.

Future research can build on the findings of this study in several directions. One focus is the improvement of pretraining and fine‑tuning strategies to enable more effective integration of multimodal features while reducing the computational cost of model adaptation.

Another area concerns the quality of input data: improved image preprocessing, both for image‑based pipelines and prior to OCR extraction, can enhance the reliability of downstream predictions. Further progress will also depend on the development of models that generalize more consistently across different layout styles, document formats and languages. In addition, support for long‑sequence and multi‑page processing should be strengthened, as realistic VRDs often contain information distributed across multiple pages or require high‑resolution inputs. Finally, the development of well‑curated benchmark datasets with clearly separated document classes would enable more stable learning and provide a more reliable foundation for evaluating multimodal models.

Beyond type classification, accurate multimodal feature-modeling is essential for downstream VRDU tasks such as document layout analysis, information extraction, and question answering. Improvements in document type classification directly benefit these areas, as advances in multimodal representation learning strengthen both the methodological foundation and the practical applicability of document understanding pipelines.

%% file: literature.bib
@misc{Ding.2025,
 author = {Ding, Yihao and Luo, Siwen and Dai, Yue and Jiang, Yanbei and Li, Zechuan and Martin, Geoffrey and Peng, Yifan},
 date = {2025},
 title = {A Survey on MLLM-based Visually Rich Document Understanding: Methods,  Challenges, and Emerging Trends},
 url = {https://arxiv.org/pdf/2507.09861}
}

@article{Ding.2026,
	author = {Ding, Yihao and Han, Soyeon Caren and Lee, Jean and Hovy, Eduard},
	year = {2026},
	title = {Deep learning based visually rich document content understanding: a survey},
	pages = {114},
	volume = {59},
	number = {4},
	issn = {1573-7462},
	journal = {Artificial Intelligence Review},
	doi = {10.1007/s10462-025-11477-3}
}

@inproceedings{LeipengHao.2016,
 author = {Hao, Leipeng and Gao, Liangcai and Yi, Xiaohan and Tang, Zhi},
 title = {A Table Detection Method for PDF Documents Based on Convolutional Neural Networks},
 pages = {287--292},
 publisher = {IEEE},
 isbn = {978-1-5090-1792-8},
 booktitle = {2016 12th IAPR Workshop on Document Analysis Systems (DAS)},
 year = {2016},
 doi = {10.1109/DAS.2016.23}
}

@misc{He.2017,
 author = {He, Kaiming and Gkioxari, Georgia and Doll{\'a}r, Piotr and Girshick, Ross},
 date = {2017},
 title = {Mask R-CNN},
 url = {https://arxiv.org/pdf/1703.06870}
}

@inproceedings{Ren.2015,
 author = {Ren, Shaoqing and He, Kaiming and Girshick, Ross and Sun, Jian},
 title = {Faster R-CNN: Towards Real-Time Object Detection with Region Proposal Networks},
 url = {https://proceedings.neurips.cc/paper_files/paper/2015/file/14bfa6bb14875e45bba028a21ed38046-Paper.pdf},
 volume = {28},
 publisher = {{Curran Associates, Inc}},
 editor = {Cortes, C. and Lawrence, N. and Lee, D. and Sugiyama, M. and Garnett, R.},
 booktitle = {Advances in Neural Information Processing Systems},
 year = {2015}
}

@inproceedings{Devlin.2019,
 author = {Devlin, Jacob and Chang, Ming-Wei and Lee, Kenton and Toutanova, Kristina},
 title = {BERT: Pre-training of Deep Bidirectional Transformers for Language Understanding},
 url = {https://aclanthology.org/N19-1423/},
 pages = {4171--4186},
 publisher = {{Association for Computational Linguistics}},
 editor = {Burstein, Jill and Doran, Christy and Solorio, Thamar},
 booktitle = {Proceedings of the 2019 Conference of the North American Chapter of the Association for Computational Linguistics: Human Language Technologies, Volume 1 (Long and Short Papers)},
 year = {2019},
 address = {Minneapolis, Minnesota},
 doi = {10.18653/v1/N19-1423}
}

@misc{Liu.2019,
 author = {Liu, Yinhan and Ott, Myle and Goyal, Naman and Du, Jingfei and Joshi, Mandar and Chen, Danqi and Levy, Omer and Lewis, Mike and Zettlemoyer, Luke and Stoyanov, Veselin},
 date = {2019},
 title = {RoBERTa: A Robustly Optimized BERT Pretraining Approach},
 url = {https://arxiv.org/pdf/1907.11692}
}

@article{Rombach.2025,
 author = {Rombach, Alexander Michael and Fettke, Peter},
 year = {2025},
 title = {Deep Learning Based Key Information Extraction from Business Documents: Systematic Literature Review},
 volume = {58},
 number = {2},
 journal = {ACM Computing Surveys},
 doi = {10.1145/3749369}
}

@inproceedings{Huang.2022,
 author = {Huang, Yupan and Lv, Tengchao and Cui, Lei and Lu, Yutong and Wei, Furu},
 title = {LayoutLMv3: Pre-training for Document AI with Unified Text and Image Masking},
 url = {http://arxiv.org/pdf/2204.08387},
 pages = {4083--4091},
 publisher = {{Association for Computing Machinery}},
 isbn = {9781450392037},
 series = {ACM Digital Library},
 editor = {Magalh{\~a}es, Jo{\~a}o},
 booktitle = {Proceedings of the 30th ACM International Conference on Multimedia},
 year = {2022},
 address = {New York, NY, United States},
 doi = {10.1145/3503161.3548112}
}

@inproceedings{Kim.2021,
 author = {Kim, Geewook and Hong, Teakgyu and Yim, Moonbin and Nam, Jeongyeon and Park, Jinyoung and Yim, Jinyeong and Hwang, Wonseok and Yun, Sangdoo and Han, Dongyoon and Park, Seunghyun},
 title = {OCR-free Document Understanding Transformer},
 url = {http://arxiv.org/pdf/2111.15664},
 pages = {498--517},
 publisher = {{Springer Nature Switzerland}},
 isbn = {978-3-031-19815-1},
 editor = {Avidan, Shai and Brostow, Gabriel and Ciss{\'e}, Moustapha and Farinella, Giovanni Maria and Hassner, Tal},
 booktitle = {Computer Vision -- ECCV 2022},
 year = {2022},
 address = {Cham}
}

@misc{Bai.2025,
 author = {Bai, Shuai and Cai, Yuxuan and Chen, Ruizhe and Chen, Keqin and Chen, Xionghui and Cheng, Zesen and Deng, Lianghao and Ding, Wei and Gao, Chang and Ge, Chunjiang and Ge, Wenbin and Guo, Zhifang and Huang, Qidong and Huang, Jie and Huang, Fei and Hui, Binyuan and Jiang, Shutong and Li, Zhaohai and Li, Mingsheng and Li, Mei and Li, Kaixin and Lin, Zicheng and Lin, Junyang and Liu, Xuejing and Liu, Jiawei and Liu, Chenglong and Liu, Yang and Liu, Dayiheng and Liu, Shixuan and Lu, Dunjie and Luo, Ruilin and Lv, Chenxu and Men, Rui and Meng, Lingchen and Ren, Xuancheng and Ren, Xingzhang and Song, Sibo and Sun, Yuchong and Tang, Jun and Tu, Jianhong and Wan, Jianqiang and Wang, Peng and Wang, Pengfei and Wang, Qiuyue and Wang, Yuxuan and Xie, Tianbao and Xu, Yiheng and Xu, Haiyang and Xu, Jin and Yang, Zhibo and Yang, Mingkun and Yang, Jianxin and Yang, An and Yu, Bowen and Zhang, Fei and Zhang, Hang and Zhang, Xi and Zheng, Bo and Zhong, Humen and Zhou, Jingren and Zhou, Fan and Zhou, Jing and Zhu, Yuanzhi and Zhu, Ke},
 date = {2025},
 title = {Qwen3-VL Technical Report},
 url = {https://arxiv.org/pdf/2511.21631}
}

@misc{Hong.2025,
 author = {Hong, Wenyi and Yu, Wenmeng and Gu, Xiaotao and Wang, Guo and Gan, Guobing and Tang, Haomiao and Cheng, Jiale and Qi, Ji and Ji, Junhui and Pan, Lihang and Duan, Shuaiqi and Wang, Weihan and Wang, Yan and Cheng, Yean and He, Zehai and Su, Zhe and Yang, Zhen and Pan, Ziyang and Zeng, Aohan and Wang, Baoxu and Chen, Bin and Shi, Boyan and Pang, Changyu and Zhang, Chenhui and {Da Yin} and Yang, Fan and Chen, Guoqing and Li, Haochen and Zhu, Jiale and Chen, Jiali and Xu, Jiaxing and Xu, Jiazheng and Chen, Jing and Lin, Jinghao and Chen, Jinhao and Wang, Jinjiang and Chen, Junjie and Lei, Leqi and Gong, Letian and Pan, Leyi and Liu, Mingdao and Xu, Mingde and Zhang, Mingzhi and Zheng, Qinkai and Lyu, Ruiliang and Tu, Shangqin and Yang, Sheng and Meng, Shengbiao and Zhong, Shi and Huang, Shiyu and Zhao, Shuyuan and Xue, Siyan and Zhang, Tianshu and Luo, Tianwei and Hao, Tianxiang and Tong, Tianyu and Jia, Wei and Li, Wenkai and Liu, Xiao and Zhang, Xiaohan and Lyu, Xin and Zhang, Xinyu and Fan, Xinyue and Huang, Xuancheng and Xue, Yadong and Wang, Yanfeng and Wang, Yanling and Wang, Yanzi and An, Yifan and Du, Yifan and Huang, Yiheng and Niu, Yilin and Shi, Yiming and Wang, Yu and Wang, Yuan and Yue, Yuanchang and Li, Yuchen and Liu, Yusen and Zhang, Yutao and Wang, Yuting and Zhang, Yuxuan and Xue, Zhao and Du, Zhengxiao and Hou, Zhenyu and Wang, Zihan and Zhang, Peng and Liu, Debing and Xu, Bin and Li, Juanzi and Huang, Minlie and Dong, Yuxiao and Tang, Jie},
 date = {2025},
 title = {GLM-4.5V and GLM-4.1V-Thinking: Towards Versatile Multimodal Reasoning with Scalable Reinforcement Learning},
 url = {https://arxiv.org/pdf/2507.01006}
}

@article{Gbada.2025,
 author = {Gbada, Hamza and Kalti, Karim and Mahjoub, Mohamed Ali},
 year = {2025},
 title = {Deep learning approaches for information extraction from visually rich documents: datasets, challenges and methods},
 pages = {121--142},
 volume = {28},
 number = {1},
 issn = {1433-2825},
 journal = {International Journal on Document Analysis and Recognition (IJDAR)},
 doi = {10.1007/s10032-024-00493-8}
}

@inproceedings{Sassioui.2023,
 author = {Sassioui, Abdellatif and Benouini, Rachid and {El Ouargui}, Yasser and {El Kamili}, Mohamed and Chergui, Meriyem and Ouzzif, Mohammed},
 title = {Visually-Rich Document Understanding: Concepts, Taxonomy and Challenges},
 pages = {1--7},
 booktitle = {2023 10th International Conference on Wireless Networks and Mobile Communications (WINCOM)},
 year = {2023},
 doi = {10.1109/WINCOM59760.2023.10322990}
}

@incollection{SciusBertrand.2024,
 author = {Scius-Bertrand, Anna and Fakhari, Atefeh and V{\"o}gtlin, Lars and Cabral, Daniel Ribeiro and Fischer, Andreas},
 title = {Are Layout Analysis and OCR Still Useful for Document Information Extraction Using Foundation Models?},
 pages = {175--191},
 volume = {14807},
 publisher = {Springer},
 isbn = {978-3-031-70545-8},
 series = {Lecture Notes in Computer Science},
 editor = {Smith, Elisa Barney and Liwicki, Marcus and Peng, Liangrui},
 booktitle = {Document analysis and recognition - ICDAR 2024},
 year = {2024},
 address = {Cham},
 doi = {10.1007/978-3-031-70546-5{\textunderscore }11}
}

@inproceedings{Borchmann.2021,
 author = {Borchmann, {\L}ukasz and Pietruszka, Micha{\l} and Stanislawek, Tomasz and Jurkiewicz, Dawid and Turski, Micha{\l} and Szyndler, Karolina and Grali{\'n}ski, Filip},
 title = {DUE: End-to-End Document Understanding Benchmark},
 url = {https://openreview.net/forum?id=rNs2FvJGDK},
 booktitle = {Thirty-fifth Conference on Neural Information Processing Systems (NeurIPS 2021) Datasets and Benchmarks Track (Round 2)},
 year = {2021}
}

@inproceedings{Larson.2023,
 author = {Larson, Stefan and Lim, Gordon and Leach, Kevin},
 title = {On Evaluation of Document Classification with RVL-CDIP},
 url = {https://aclanthology.org/2023.eacl-main.195/},
 pages = {2665--2678},
 publisher = {{Association for Computational Linguistics}},
 editor = {Vlachos, Andreas and Augenstein, Isabelle},
 booktitle = {Proceedings of the 17th Conference of the European Chapter of the Association for Computational Linguistics},
 year = {2023},
 address = {Dubrovnik, Croatia},
 doi = {10.18653/v1/2023.eacl-main.195}
}

@misc{Yang.2025,
 author = {Bao, Keqin and Cui, Zeyu and Dang, Kai and Deng, Lianghao and Fan, Yang and Gao, Ruize and Gao, Chang and Ge, Hao and Hu, Feng and Huang, Chengen and Huang, Fei and Hui, Binyuan and {Le Yu} and Li, Anfeng and Li, Mingze and Li, Mei and Li, Tianhao and Lin, Huan and Lin, Junyang and Liu, Dayiheng and Liu, Shixuan and Liu, Yuqiong and Luo, Shuang and Lv, Chenxu and Men, Rui and Qiu, Zihan and Ren, Xingzhang and Ren, Xuancheng and Su, Yang and Tang, Jialong and Tang, Tianyi and Tu, Jianhong and Wan, Yu and Wang, Xinyu and Wang, Peng and Wang, Zekun and Wei, Haoran and Xue, Mingfeng and Yang, Kexin and Yang, An and Yang, Baosong and Yang, Jiaxi and Yang, Jianxin and Yang, Jian and Yin, Wenbiao and Yu, Bowen and Zhang, Zhenru and Zhang, Beichen and Zhang, Yinger and Zhang, Yichang and Zhang, Pei and Zhang, Xinyu and Zhang, Jianwei and Zheng, Bo and Zheng, Chujie and Zhou, Jing and Zhou, Jingren and Zhou, Zhipeng and Zhou, Fan and Zhu, Qin},
 date = {2025},
 title = {Qwen3 Technical Report},
 url = {https://arxiv.org/pdf/2505.09388}
}

@inproceedings{Harley.2015,
 author = {Harley, Adam W. and Ufkes, Alex and Derpanis, Konstantinos G.},
 title = {Evaluation of deep convolutional nets for document image classification and retrieval},
 pages = {991--995},
 publisher = {IEEE},
 isbn = {978-1-4799-1805-8},
 booktitle = {13th International Conference on Document Analysis and Recognition (ICDAR 2015)},
 year = {2015},
 address = {Piscataway, NJ},
 doi = {10.1109/ICDAR.2015.7333910}
}

@article{Appalaraju.2024,
 author = {Appalaraju, Srikar and Tang, Peng and Dong, Qi and Sankaran, Nishant and Zhou, Yichu and Manmatha, R.},
 year = {2024},
 title = {DocFormerv2: Local Features for Document Understanding},
 pages = {709--718},
 volume = {38},
 number = {2},
 journal = {Proceedings of the AAAI Conference on Artificial Intelligence (AAAI)},
 doi = {10.1609/aaai.v38i2.27828}
}

@misc{IDL.2026,
 author = {{University of California, San Francisco}},
 title = {Industry Documents Library},
 url = {https://www.industrydocuments.ucsf.edu/},
 urldate = {07.05.2026}
}

@inproceedings{Lewis.2006,
 author = {Lewis, D. and Agam, G. and Argamon, S. and Frieder, O. and Grossman, D. and Heard, J.},
 title = {Building a test collection for complex document information processing},
 pages = {665--666},
 publisher = {{ACM Press}},
 isbn = {1595933697},
 editor = {Dumais, Susan},
 booktitle = {Proceedings of the Twenty-Ninth Annual International ACM SIGIR Conference on Research and Development in Information Retrieval},
 year = {2006},
 address = {New York, NY},
 doi = {10.1145/1148170.1148307}
}

@inproceedings{SeunghyunPark.2019,
 author = {Park, Seunghyun and Shin, Seung and Lee, Bado and Lee, Junyeop and Surh, Jaeheung and Seo, Minjoon and Lee, Hwalsuk},
 title = {CORD: A Consolidated Receipt Dataset for Post-OCR Parsing},
 url = {https://openreview.net/forum?id=SJl3z659UH},
 booktitle = {Document Intelligence Workshop at Neural Information Processing Systems (NeurIPS)},
 year = {2019}
}

@inproceedings{Jaume.2019,
 author = {Jaume, Guillaume and Ekenel, Hazim and Thiran, Jean-Philippe},
 title = {FUNSD: A Dataset for Form Understanding in Noisy Scanned Documents},
 url = {https://arxiv.org/pdf/1905.13538},
 booktitle = {Accepted to ICDAR-OST},
 year = {2019}
}

@inproceedings{Bhattacharyya.2025,
 author = {Bhattacharyya, Aniket and Tripathi, Anurag and Das, Ujjal and Karmakar, Archan and Pathak, Amit and Gupta, Maneesh},
 title = {Information Extraction from Visually Rich Documents using LLM-based Organization of Documents into Independent Textual Segments},
 pages = {17241--17256},
 publisher = {{Association for Computational Linguistics}},
 editor = {Che, Wanxiang and Nabende, Joyce and Shutova, Ekaterina and Pilehvar, Mohammad Taher},
 booktitle = {Proceedings of the 63rd Annual Meeting of the Association for Computational Linguistics (Volume 1: Long Papers)},
 year = {2025},
 address = {Stroudsburg, PA, USA},
 doi = {10.18653/v1/2025.acl-long.844}
}

@inproceedings{Luo.2024,
 author = {Luo, Chuwei and Shen, Yufan and Zhu, Zhaoqing and Zheng, Qi and Yu, Zhi and Yao, Cong},
 title = {LayoutLLM: Layout Instruction Tuning with Large Language Models for Document Understanding},
 pages = {15630--15640},
 booktitle = {2024 IEEE/CVF Conference on Computer Vision and Pattern Recognition (CVPR)},
 year = {2024},
 doi = {10.1109/CVPR52733.2024.01480}
}

@inproceedings{Ye.2023,
 author = {Ye, Jiabo and Hu, Anwen and Xu, Haiyang and Ye, Qinghao and Yan, Ming and Xu, Guohai and Li, Chenliang and Tian, Junfeng and Qian, Qi and Zhang, Ji and Jin, Qin and He, Liang and Lin, Xin and Huang, Fei},
 title = {UReader: Universal OCR-free Visually-situated Language Understanding with Multimodal Large Language Model},
 url = {https://aclanthology.org/2023.findings-emnlp.187/},
 pages = {2841--2858},
 publisher = {{Association for Computational Linguistics}},
 editor = {Bouamor, Houda and Pino, Juan and Bali, Kalika},
 booktitle = {Findings of the Association for Computational Linguistics: EMNLP 2023},
 year = {2023},
 address = {Singapore},
 doi = {10.18653/v1/2023.findings-emnlp.187}
}

@inproceedings{Jaaskelainen.2023,
 author = {J{\"a}{\"a}skel{\"a}inen, Anssi and Lipsanen, Mikko and F{\"o}hr, Atte and R{\"a}is{\"a}nen, Tuomo},
 title = {OCR quality: Key to enhanced Data Mining},
 pages = {1--6},
 booktitle = {2023 3rd International Conference on Electrical, Computer, Communications and Mechatronics Engineering (ICECCME)},
 year = {2023},
 doi = {10.1109/ICECCME57830.2023.10252214}
}

@article{SouhailBakkali.2023,
 author = {{Souhail Bakkali} and {Zuheng Ming} and {Mickael Coustaty} and {Mar{\c{c}}al Rusi{\~n}ol} and {Oriol Ramos Terrades}},
 year = {2023},
 title = {VLCDoC: Vision-Language contrastive pre-training model for cross-Modal document classification},
 url = {https://www.sciencedirect.com/science/article/pii/S0031320323001206},
 pages = {109419},
 volume = {139},
 issn = {0031-3203},
 journal = {Pattern Recognition},
 doi = {10.1016/j.patcog.2023.109419}
}

@incollection{Skalicky.2022,
 author = {Skalick{\'y}, Maty{\'a}{\v{s}} and {\v{S}}imsa, {\v{S}}t{\v{e}}p{\'a}n and U{\v{r}}i{\v{c}}{\'a}{\v{r}}, Michal and {\v{S}}ulc, Milan},
 title = {Business Document Information Extraction: Towards Practical Benchmarks},
 pages = {105--117},
 volume = {13390},
 publisher = {{Springer International Publishing} and {Imprint Springer}},
 isbn = {978-3-031-13642-9},
 series = {Lecture Notes in Computer Science},
 editor = {Barr{\'o}n-Cede{\~n}o, Alberto and {Da San Martino}, Giovanni and {Degli Esposti}, Mirko and Sebastiani, Fabrizio and Macdonald, Craig and Pasi, Gabriella and Hanbury, Allan and Potthast, Martin and Faggioli, Guglielmo and Ferro, Nicola},
 booktitle = {Experimental IR Meets Multilinguality, Multimodality, and Interaction},
 year = {2022},
 address = {Cham},
 doi = {10.1007/978-3-031-13643-6{\textunderscore }8}
}

@inproceedings{Pfitzmann.2022,
 author = {Pfitzmann, Birgit and Auer, Christoph and Dolfi, Michele and Nassar, Ahmed S. and Staar, Peter},
 title = {DocLayNet: A Large Human-Annotated Dataset for Document-Layout Segmentation},
 pages = {3743--3751},
 publisher = {{Association for Computing Machinery}},
 isbn = {9781450393850},
 series = {ACM Digital Library},
 editor = {Zhang, Aidong},
 booktitle = {Proceedings of the 28th ACM SIGKDD Conference on Knowledge Discovery and Data Mining},
 year = {2022},
 address = {New York, United States},
 doi = {10.1145/3534678.3539043}
}

@article{Abdallah.2024,
 author = {Abdallah, Abdelrahman and Eberharter, Daniel and Pfister, Zoe and Jatowt, Adam},
 year = {2024},
 title = {A survey of recent approaches to form understanding in scanned documents},
 pages = {342},
 volume = {57},
 number = {12},
 issn = {1573-7462},
 journal = {Artificial Intelligence Review},
 doi = {10.1007/s10462-024-11000-0}
}

@inproceedings{Xu.2022,
 author = {Xu, Yiheng and Lv, Tengchao and Cui, Lei and Wang, Guoxin and Lu, Yijuan and Florencio, Dinei and Zhang, Cha and Wei, Furu},
 title = {XFUND: A Benchmark Dataset for Multilingual Visually Rich Form Understanding},
 url = {https://aclanthology.org/2022.findings-acl.253/},
 pages = {3214--3224},
 publisher = {{Association for Computational Linguistics}},
 editor = {Muresan, Smaranda and Nakov, Preslav and Villavicencio, Aline},
 booktitle = {Findings of the Association for Computational Linguistics: ACL 2022},
 year = {2022},
 address = {Dublin, Ireland},
 doi = {10.18653/v1/2022.findings-acl.253}
}

@inproceedings{Liu.2021,
 author = {Liu, Ze and Lin, Yutong and Cao, Yue and Hu, Han and Wei, Yixuan and Zhang, Zheng and Lin, Stephen and Guo, Baining},
 title = {Swin Transformer: Hierarchical Vision Transformer using Shifted Windows},
 pages = {9992--10002},
 booktitle = {2021 IEEE/CVF International Conference on Computer Vision (ICCV)},
 year = {2021},
 doi = {10.1109/ICCV48922.2021.00986}
}

@inproceedings{Lewis.2020,
 author = {Lewis, Mike and Liu, Yinhan and Goyal, Naman and Ghazvininejad, Marjan and Mohamed, Abdelrahman and Levy, Omer and Stoyanov, Veselin and Zettlemoyer, Luke},
 title = {BART: Denoising Sequence-to-Sequence Pre-training for Natural Language Generation, Translation, and Comprehension},
 url = {https://aclanthology.org/2020.acl-main.703/},
 pages = {7871--7880},
 publisher = {{Association for Computational Linguistics}},
 editor = {Jurafsky, Dan and Chai, Joyce and Schluter, Natalie and Tetreault, Joel},
 booktitle = {Proceedings of the 58th Annual Meeting of the Association for Computational Linguistics},
 year = {2020},
 address = {Online},
 doi = {10.18653/v1/2020.acl-main.703}
}

@article{Liu.2020,
 author = {Liu, Yinhan and Gu, Jiatao and Goyal, Naman and Li, Xian and Edunov, Sergey and Ghazvininejad, Marjan and Lewis, Mike and Zettlemoyer, Luke and Johnson, Mark and Roark, Brian and Nenkova, Ani},
 year = {2020},
 title = {Multilingual Denoising Pre-training for Neural Machine Translation},
 url = {https://aclanthology.org/2020.tacl-1.47/},
 pages = {726--742},
 volume = {8},
 journal = {Transactions of the Association for Computational Linguistics},
 doi = {10.1162/tacl{\textunderscore }a{\textunderscore }00343}
}

@inproceedings{Li.2022,
 author = {Li, Junlong and Xu, Yiheng and Lv, Tengchao and Cui, Lei and Zhang, Cha and Wei, Furu},
 title = {DiT: Self-supervised Pre-training for Document Image Transformer},
 pages = {3530--3539},
 publisher = {{Association for Computing Machinery}},
 isbn = {9781450392037},
 series = {ACM Digital Library},
 editor = {Magalh{\~a}es, Jo{\~a}o},
 booktitle = {Proceedings of the 30th ACM International Conference on Multimedia},
 year = {2022},
 address = {New York, NY, United States},
 doi = {10.1145/3503161.3547911}
}

@misc{Tschannen.2025,
 author = {Tschannen, Michael and Gritsenko, Alexey and Wang, Xiao and Naeem, Muhammad Ferjad and Alabdulmohsin, Ibrahim and Parthasarathy, Nikhil and Evans, Talfan and Beyer, Lucas and Xia, Ye and Mustafa, Basil and H{\'e}naff, Olivier and Harmsen, Jeremiah and Steiner, Andreas and Zhai, Xiaohua},
 date = {2025},
 title = {SigLIP 2: Multilingual Vision-Language Encoders with Improved Semantic Understanding, Localization, and Dense Features},
 url = {https://arxiv.org/pdf/2502.14786}
}

@misc{Yang.2024,
 author = {Yang, An and Yang, Baosong and Hui, Binyuan and Zheng, Bo and Yu, Bowen and Zhou, Chang and Li, Chengpeng and Li, Chengyuan and Liu, Dayiheng and Huang, Fei and Dong, Guanting and Wei, Haoran and Lin, Huan and Tang, Jialong and Wang, Jialin and Yang, Jian and Tu, Jianhong and Zhang, Jianwei and Ma, Jianxin and Yang, Jianxin and Xu, Jin and Zhou, Jingren and Bai, Jinze and He, Jinzheng and Lin, Junyang and Dang, Kai and Lu, Keming and Chen, Keqin and Yang, Kexin and Li, Mei and Xue, Mingfeng and Ni, Na and Zhang, Pei and Wang, Peng and Peng, Ru and Men, Rui and Gao, Ruize and Lin, Runji and Wang, Shijie and Bai, Shuai and Tan, Sinan and Zhu, Tianhang and Li, Tianhao and Liu, Tianyu and Ge, Wenbin and Deng, Xiaodong and Zhou, Xiaohuan and Ren, Xingzhang and Zhang, Xinyu and Wei, Xipin and Ren, Xuancheng and Liu, Xuejing and Fan, Yang and Yao, Yang and Zhang, Yichang and Wan, Yu and Chu, Yunfei and Liu, Yuqiong and Cui, Zeyu and Zhang, Zhenru and Guo, Zhifang and Fan, Zhihao},
 date = {2024},
 title = {Qwen2 Technical Report},
 url = {https://arxiv.org/pdf/2407.10671}
}

@inproceedings{Wolf.2020,
 author = {Wolf, Thomas and Debut, Lysandre and Sanh, Victor and Chaumond, Julien and Delangue, Clement and Moi, Anthony and Cistac, Pierric and Rault, Tim and Louf, Remi and Funtowicz, Morgan and Davison, Joe and Shleifer, Sam and von Platen, Patrick and Ma, Clara and Jernite, Yacine and Plu, Julien and Xu, Canwen and {Le Scao}, Teven and Gugger, Sylvain and Drame, Mariama and Lhoest, Quentin and Rush, Alexander},
 title = {HuggingFace Transformers: State-of-the-Art Natural Language Processing},
 url = {https://aclanthology.org/2020.emnlp-demos.6/},
 pages = {38--45},
 publisher = {{Association for Computational Linguistics}},
 editor = {Liu, Qun and Schlangen, David},
 booktitle = {Proceedings of the 2020 Conference on Empirical Methods in Natural Language Processing: System Demonstrations},
 year = {2020},
 address = {Online},
 doi = {10.18653/v1/2020.emnlp-demos.6}
}
